\documentclass{article}

\PassOptionsToPackage{numbers, compress}{natbib}

\usepackage[preprint]{neurips_2023}

\usepackage[utf8]{inputenc} 
\usepackage[T1]{fontenc}    
\usepackage{url}            
\usepackage{amsfonts}       
\usepackage{nicefrac}       
\usepackage{microtype}      
\usepackage{xcolor}         

\usepackage{xspace}
\usepackage{graphicx}
\usepackage{amsmath}
\usepackage{amssymb}
\usepackage{wrapfig}
\usepackage{caption}
\usepackage{subcaption}
\usepackage{adjustbox}
\usepackage{multirow,booktabs}
\usepackage{colortbl}
\usepackage{tcolorbox}
\usepackage{enumitem}
\usepackage{makecell}
\usepackage{listings}
\usepackage{algorithm,algorithmic}
\usepackage{setspace}
\usepackage[accsupp]{axessibility} 
\usepackage{bbding}
\usepackage{etoc}
\usepackage{enumitem}
\etocdepthtag.toc{mtchapter}
\etocsettagdepth{mtchapter}{subsection}
\etocsettagdepth{mtappendix}{none}

\definecolor{darkblue}{rgb}{0, 0.12, 0.55}
\definecolor{darkgreen}{rgb}{0, 0.55, 0.12}
\definecolor{darkred}{rgb}{0.6,0,0}
\definecolor{darkgreen}{rgb}{0,0.6,0}
\definecolor{Gray}{gray}{0.9}
\definecolor{mygreen}{HTML}{3cb44b}
\newcommand{\PredSty}[1]{\textnormal{\ttfamily\color{mygreen!90!black}#1}\unskip}
\usepackage[breaklinks=true,
            colorlinks,
            linkcolor = darkred,
            urlcolor  = darkblue, 
            citecolor = teal,
            bookmarks = false]{hyperref}

\usepackage[capitalize]{cleveref}
\Crefname{section}{Section}{Section}
\Crefname{table}{Table}{Tables}
\Crefname{figure}{Figure}{Figure}

\newcommand{\llama}{Llama}
\newcommand{\vicuna}{Vicuna}

\newcommand{\model}{MotionLLM\xspace}
\newcommand{\data}{MoVid\xspace}
\newcommand{\bench}{MoVid-Bench\xspace}

\definecolor{mylinkcolor}{RGB}{0,0,0}

\author{Ling-Hao Chen\textbf{\thanks{Equal contribution, random listing order. Work done by Ling-Hao Chen, Shunlin Lu, and Hao Zhang during internship at IDEA Research.} \ $^{1, 3}$}, Shunlin Lu$^{*2, 3}$, \\ \textbf{Ailing Zeng$^{3}$, Hao Zhang$^{3, 4}$, Benyou Wang$^{2}$, Ruimao Zhang$^{2}$, Lei Zhang\thanks{Correspondence: Lei Zhang.} \ $^{3}$}
\\
\texttt{\{thu.lhchen, shunlinlu0803\}@gmail.com}
\\
$^{1}$Tsinghua University \\ $^{2}$School of Data Science, The Chinese University of Hong Kong, Shenzhen (CUHK-SZ) \\ $^{3}$International Digital Economy Academy (IDEA Research) \\ $^{4}$The Hong Kong University of Science and Technology\\
Project page: \url{https://lhchen.top/MotionLLM}
}

\begin{document}

\newcommand{\logo}{\parbox[c]{.06\linewidth}{\includegraphics[width=\linewidth]{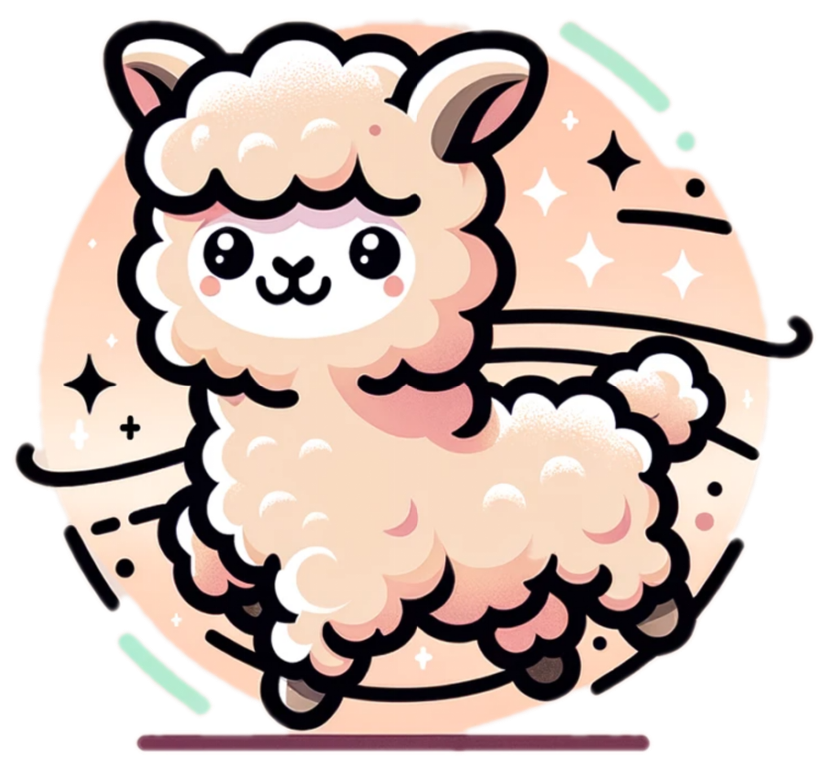}}}
\newcommand{\smalllogo}{\parbox[c]{.04\linewidth}{\includegraphics[width=\linewidth]{figure/logo.png}}}

\title{\logo \ MotionLLM: Understanding Human Behaviors from Human Motions and Videos}


\maketitle
\vspace{-2em}

\begin{figure}[!h]
    \centering     
    \begin{subfigure}{\textwidth}
      \centering   
      \includegraphics[width=1\linewidth]{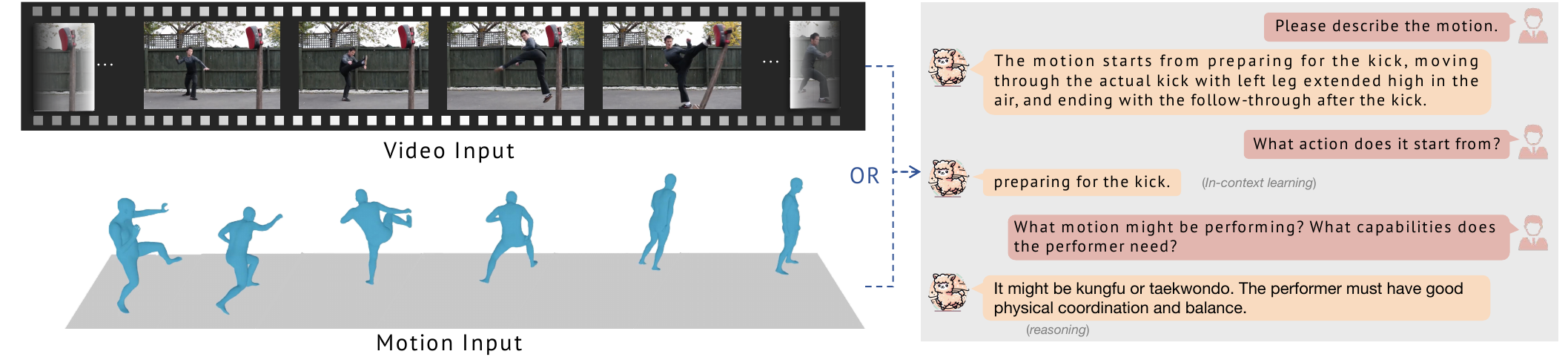}
      \vspace{-1.6em}
        \caption{Our MotionLLM takes motions or videos as inputs to understand human behaviors.}
        \label{fig:tesear1}
    \end{subfigure} 
    \begin{subfigure}{\textwidth}
      \centering   
      \includegraphics[width=1\linewidth]{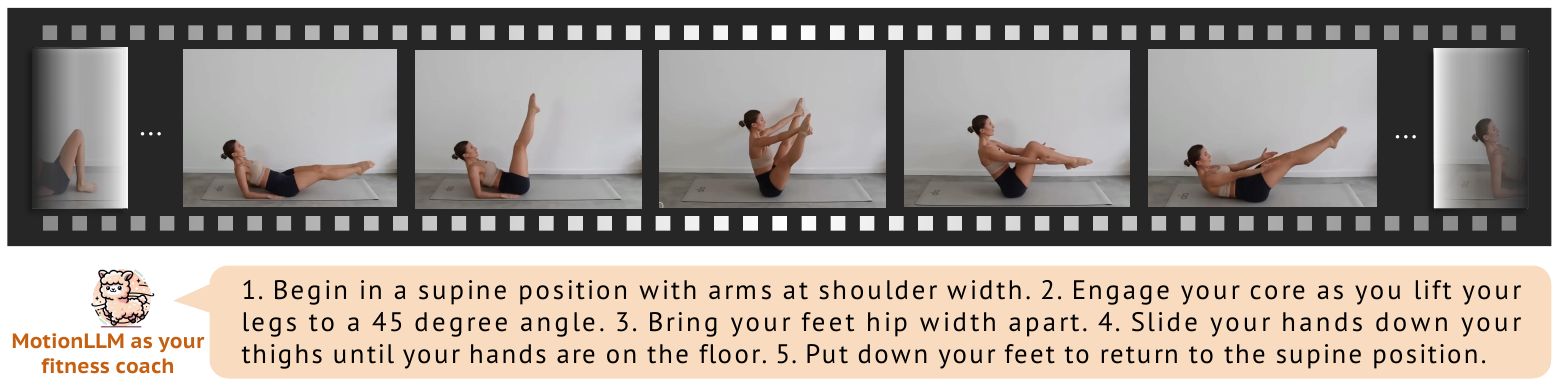}
      \vspace{-1.8em}
        \caption{An application: MotionLLM as your fitness coach based on its caption capability.}
        \label{fig:tesear2}
    \end{subfigure} 
    \vspace{-1.8em}
    \caption{{\bf Introducting MotionLLM.} (a) The input and output of MotionLLM. (b) MotionLLM has broad application scenarios, such as an intelligent fitness coach.}
    \label{fig:tesear}
    \vspace{-0.2em}
\end{figure}

\begin{abstract}
\vspace{-0.4em}
This study delves into the realm of multi-modality (\emph{i.e.}, video and motion modalities) human behavior understanding by leveraging the powerful capabilities of Large Language Models (LLMs). Diverging from recent LLMs designed for video-only or motion-only understanding, we argue that understanding human behavior necessitates joint modeling from both videos and motion sequences (\emph{e.g.}, SMPL sequences) to capture nuanced body part dynamics and semantics effectively. In light of this, we present MotionLLM, a straightforward yet effective framework for human motion understanding, captioning, and reasoning. Specifically, MotionLLM adopts a unified video-motion training strategy that leverages the complementary advantages of existing coarse video-text data and fine-grained motion-text data to glean rich spatial-temporal insights. Furthermore, we collect a substantial dataset, MoVid, comprising diverse videos, motions, captions, and instructions. Additionally, we propose the MoVid-Bench, with carefully manual annotations, for better evaluation of human behavior understanding on video and motion. Extensive experiments show the superiority of MotionLLM in the caption, spatial-temporal comprehension, and reasoning ability.
\end{abstract}

\etocdepthtag.toc{mtappendix}
\etocsettagdepth{mtchapter}{none}
\etocsettagdepth{mtappendix}{subsection}

\definecolor{cvprblue}{rgb}{0.21,0.49,0.74}
{
\normalsize
\begingroup
\hypersetup{linkcolor=cvprblue}
\tableofcontents

\endgroup
}
\newpage

\section{Introduction}
\label{sec:intro} 

Understanding human behavior, such as fine-grained captioning and analysis, is crucial in the realm of human-centric multi-modality intelligence~\cite{hong2022versatile,motiongpt,avatargpt} and can benefit embodied intelligence from human-computer interaction and robotics to healthcare and security~\cite{wang2022towards,wang2023learning,pan2023synthesizing,xiao2023unified,diffhoi,peng2023hoi,li2023dfa3d,li2022dcl}.
Recently, there has been notable progress in general-purpose visual understanding~\cite{videollava,zhang2023video,chen2023videollm,jin2023chat,maaz2023video,wang2023gpt4video,wang2023chatvideo,zhang2023bev,lin2023mm}, owing to the emergence of Large Language Models~(LLMs)~\cite{chatgpt,llama,vicuna}. Nevertheless, there still remains a significant gap in obtaining a good understanding of human behaviors on spatial-temporal dynamics, fine-grained semantics, behavior reasoning, \etc.

Human behaviors can be mainly represented by extracted human motions (\eg, via 3D human parametric model SMPL~\cite{SMPL:2015,amass} or skeleton sequences ~\cite{shahroudy2016ntu}) or videos~\cite{yue2022action,murray2012ava,yu2019activitynet,caba2015activitynet}. Although extracted human motion is a kind of low-redundant, appearance-invariance, and privacy-friendly representation, directly obtaining high-quality motions needs expensive motion-capture processes~\cite{vlasic2007practical,bodenheimer1997process,moeslund2006survey,amass}, resulting in the scarcity. Besides, deficiencies in the motion-environment interaction of motion data will lead to insufficient understanding of behaviors.
In contrast, human-centric videos are easy to obtain and contain rich human-environment interaction visual cues, which helps semantic motion understanding and reasoning holistically. For example, \textit{playing golf} and \textit{sweeping the floor} are similar motions but are quite different in video contexts. However, videos inevitably have high computation costs, raise privacy concerns, and contain excessively redundant elements and ambiguities instead of focusing on humans.

Considering the complementary combination of compact motions and rich-context videos, we argue that jointly modeling them is essential to pursue a more accurate, robust, and contextually rich understanding of the dynamics and semantics of motions.
Nevertheless, existing works either use motions~\cite{plappert2018learning,yamada2018paired,tm2t,takano2015statistical,motiongpt,avatargpt} or videos~\cite{videollava,zhang2023video,chen2023videollm,jin2023chat,maaz2023video,wang2023gpt4video,wang2023chatvideo,lin2023mm,song2023moviechat,videochatcaptioner,pan2023scanning,yuen2023merlin} as inputs separately to conduct human-centric motion or action understanding with LLMs. 
We attribute the challenges of this problem to two critical points: 1) limited high-quality video-motion-text pairs and the instruction tuning data; 2) under-explored problem to integrate motion and video understanding into a unified system due to lack of data and incomplete harmonization among text, motion, and video modalities.

To address the aforementioned challenges, this work attempts to lay the foundation of human-centric motion-video-text paired data and a unified understanding framework. \textbf{Firstly}, we introduce the \data dataset, comprising diverse videos, motions, captions, and instructions. The texts contain captions and instruction question-answers (QAs) to support different tasks and training stages. Motion data is sourced from existing large-scale datasets, including AMASS~\cite{amass} (captions available from HumanML3D~\cite{guo2022generating}) and Motion-X~\cite{motionx} (accompanied with videos). Regarding video-text data, we use GPT-4V~\cite{gpt4,wu2023early,yang2023dawn} to annotate 24k video captions from Motion-X, employing a 15$\times$ down-sampling rate for keyframes alongside meticulously designed prompts. 
For motion-text data, we augment the manually annotated captions of HumanML3D via GPT-4~\cite{gpt4}, resulting in 272k QA pairs serving as instructions. To facilitate effective instruction tuning, these instructions encompass diverse spatial-temporal questions, in-context QA pairs, and reasoning data. Similarly, we obtain 200k instructions for Motion-X.
\textbf{Secondly}, we propose \model to understand human behaviors with motion and videos in one system (\cref{fig:tesear1}). Technically, to project motions and videos into the linguistic space via trainable motion/video translators as V-L translators in the first stage. It enables the unification of human behaviors with various modalities as translated languages, thereby leveraging the reasoning ability inherent in LLMs~\cite{vicuna}. In the second stage, we fine-tune the LLM and V-L translators through motion-video joint instruction tuning. By sharing the knowledge from both modalities in the linguistic space of LLM, \model can take advantage of the compatibility of two modalities.

For a fair and thorough evaluation of both motion and video understanding, we present a benchmark, namely \bench, to evaluate model performance on sequential dynamics, body-part semantics, direction awareness, reasoning ability, and robustness against hallucination using diverse metrics. The reference answers undergo meticulous human annotation and verification. Compared to MotionGPT~\cite{motiongpt} and Video-LLaVA~\cite{videollava}, MotionLLM demonstrates average improvements of 38\% and 15\% in motion and video understanding, respectively. In our ablation study, integrating fine-grained motion gains an average 15\% enhancement in video-based understanding, while visual content cues from videos improve motion-based understanding by 29\%. Our extensive evaluation yields valuable insights into human behavior understanding to community development and follow-up research. Lastly, empowered with superior understanding capabilities from human motions and videos, MotionLLM exhibits versatility across various downstream applications, such as serving as a fitness coach for social goods~(\cref{fig:tesear2}), particularly catering to the visually impaired community.

Before delivering into detail, we summarize our key contributions as follows.

\begin{itemize}
    \item {
        To relieve the scarcity of data issues, we introduce \data with diverse caption and instruction annotations from motion/video datasets for training holistic spatial-temporal understanding and fine-grained behaviors.
    }
    \item {
        To bridge the gap between video and motion modalities, we propose \ a model with a unified video-motion training strategy for human behavior understanding, captioning, and reasoning.
    }
    \item {
        For better evaluation of fine-grained understanding, we carefully construct a \bench benchmark considering many motion-related aspects.
    }
\end{itemize}

\section{Related Work}
\label{sec:relatedwork}

\subsection{LLM-based Video Understanding}

Video understanding plays a pivotal role in numerous applications across various domains due to its ability to extract meaningful insights and information from visual data. Previous attempts~\cite{shi2023learning,yu2018fine,li2023generating,yang2023vid2seq} try to generate captions of video content with deep learning models. The defects of these methods are mainly related to poor reasoning and understanding abilities. With the notable success of Large Language Models (LLMs), there emerges a series of vision-based or multimodal LLMs~\cite{llava,llavamed,llavagrounding,videollava,llavaplus,cogvlm} and the corresponding benchmarks~\cite{mvbench,ning2023video}. Recently, these methods~\cite{videollava,zhang2023video,chen2023videollm,jin2023chat,maaz2023video,wang2023gpt4video,wang2023chatvideo,lin2023mm,zhao2024distilling} explore the general-purpose understanding of video contents and the reasoning ability about the videos. 
Specifically, Video-LLaVA enables an LLM to perform visual reasoning capabilities on images and videos simultaneously. It learns a united visual representation by aligning images and videos before projecting them into the language feature space. 
However, due to paired data limitation and ignoring the differences in motion representations from images, there still exists a significant gap in understanding human-centric behaviors in the videos, especially dynamic movements for fine-grained body semantics.


\subsection{Human Motion Understanding}
Human motion understanding~\cite{plappert2018learning,yamada2018paired,tm2t,motiongpt,avatargpt} aims to extract the semantics of human motions. It is quite fundamental and promising for autonomous textual annotation and analysis for human motions, paving the path to build up more data for text-aligned motion generation~\cite{dai2024motionlcm,ahn2018text2action,temos,motiondiffuse,mld,guo2022generating,ahuja2019language2pose,ghosh2021synthesis,t2m-gpt,lee2023multiact,zhou2023ude,dabral2023mofusion,humantomato,guo2023momask,xie2023omnicontrol,shi2023toss,huang2024dreamwaltz,chen2023humanmac,lu2023humantomato,zhou2023emdm,peng2023hoi,liu2023interactive,shi2023generating}. Takano~\etal~\cite{takano2015statistical} takes the early attempt to generate a textual description of a motion via statistical methods. PoseScript~\cite{posescript} is proposed to describe the single-frame pose, which enjoys good performance on the spatial motion understanding but ignores the temporal motion understanding. Besides, \cite{plappert2018learning,yamada2018paired,tm2t} proposed deep models to perform motion captioning. Recently, some works~\cite{motiongpt,avatargpt} have introduced LLMs to understand human poses or motions. However, these attempts mainly focus on motion captioning and are not equipped with detailed spatial-temporal awareness and reasoning abilities. As analyzed in~\cite{motiongpt}, due to the limited motion and instruction tuning data, these works are not capable of the reasoning ability and are hard to adapt to larger LLMs, \eg \llama~\cite{llama} or \vicuna~\cite{vicuna}. Besides, the all-in-one system of motion generation and understanding is a kind of compromising unification. Instead, in MotionLLM, we project the motion and video data into the linguistic space to obtain a better understanding of motions and videos. Besides, with the reasoning ability of LLMs, we can take advantage of the compatibility on both modalities.

\section{Methodology}
\label{sec:method}

\subsection{Preliminaries and Notations}

We begin by clarifying the preliminaries and notations in MotionLLM. MotionLLM takes visual prompts $\texttt{\bf P} = \texttt{\bf M} \vee \texttt{\bf V}$ (a motion $\texttt{\bf M}$ or a video $\texttt{\bf V}$) as input, and outputs the text sequence ${\bf \texttt{z}}=\{{\bf \texttt{z}}^{1}, {\bf \texttt{z}}^{2}, \cdots, {\bf \texttt{z}}^{L}\}\in \{0, 1\}^{L\times |\mathbb{S}|}$ that follows the prompts, where $\mathbb{S}$ denotes the vocabulary set. Specifically, a motion $\texttt{\bf M}$ is composed with $F$-frame pose sequences $\texttt{\bf M} = \{{\bf m}^{1}, {\bf m}^{2}, \cdots, {\bf m}^{F} \}$ and a video composed with $T$ key-frame image sequences $\texttt{\bf V} = \{{\bf v}^{1}, {\bf v}^{2}, \cdots, {\bf v}^{T} \}$. The text generation problem can be formulated as an auto-regressive problem: ${\bf \texttt{z}} = F({\bf \texttt{z}}_{l} \mid \texttt{\bf P}, {\bf \texttt{z}}_{<\ell})$, where $F(\cdot)$ is MotionLLM. The training process of MotionLLM uses a cross-entropy loss $\mathcal{L}=-\sum_{\ell=1}^{L} F\left({\bf \texttt{z}}_{\ell} \mid \texttt{\bf P}, {\bf \texttt{z}}_{<\ell}\right)$.

\subsection{MotionLLM: Understanding Human Motions and Videos 
}
\label{sec:arch}

\begin{figure}[!t]
    \centering     
    \begin{subfigure}{0.73\textwidth}
      \centering   
      \includegraphics[width=1\linewidth]{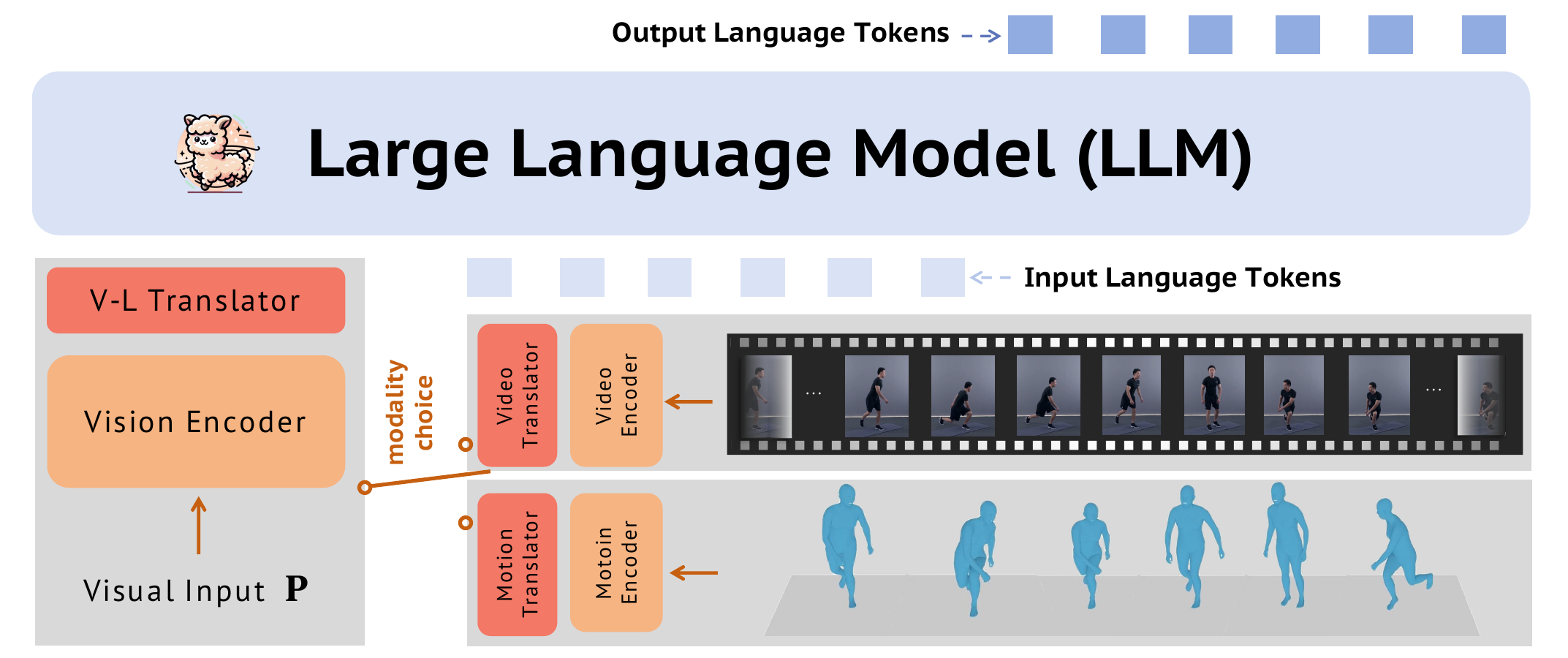}
        \caption{MotionLLM Architecture.}
        \label{fig:motionllm_s1}
    \end{subfigure} 
    \begin{subfigure}{0.23\textwidth}
      \centering   
      \includegraphics[width=0.93\linewidth]{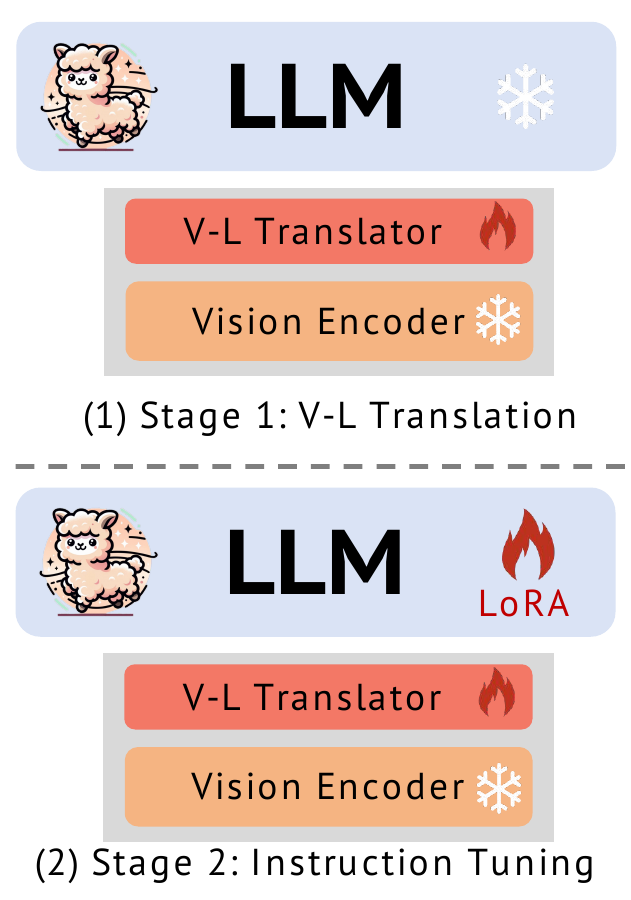}
        \caption{Two-stage  Tuning.}
        \label{fig:motionllm_s2}
    \end{subfigure} 
    \vspace{-0.7em}
    \caption{{\bf System overview of MotionLLM.} (a) MotionLLM takes videos or human motions as visual input $\mathbf{V}$. It first processes the visual input with a vision encoder and translates the vision embeddings into linguistic space via a V-L translator. (b) MotionLLM is trained in two stages. In the first stage, we train the V-L translator to learn the modality translation. In the second stage, we fine-tune the LLM and the V-L translator via instruction tuning data.
    }
    \label{fig:motionllm_system}
    \vspace{-1.5em}
\end{figure}

\noindent{\textbf{System overview.}}
%
As shown in~\cref{fig:motionllm_s1}, MotionLLM takes videos or human motions as visual prompts $\mathbf{P}$. MotionLLM first processes the visual prompts $\mathbf{P}$ with a vision encoder and translates the vision embeddings into linguistic space via a V-L translator. Note that we only take one video or motion data as input. With a well-trained MotionLLM, we output the languages in an auto-regressive fashion, \textit{i.e.} ${\bf \texttt{z}} = F({\bf \texttt{z}}_{l} \mid \texttt{\bf P}, {\bf \texttt{z}}_{<\ell})$. The training of MotionLLM can be divided into two stages. As shown in~\cref{fig:motionllm_s2}, in the first stage, MotionLLM learns a translation layer~(V-L translator, $f_{\bf T}(\cdot)$) between vision embeddings and LLM for bridging the modality gap. Here, the vision embeddings are obtained by vision encoders $f_{\bf E}(\cdot)$. In the second stage, MotionLLM fine-tunes both the V-L translator and LLM parts, \textit{i.e.} $f_{\bf L}(\cdot)$, via instruction tuning data. The whole MotionLLM can be treated as a composite function of $F=f_{\bf E}\circ f_{\bf T}\circ f_{\bf L}$. We detail both training parts as follows.

\noindent{\textbf{Modality translation~(Stage 1).}}
As there exists a modality gap between visual contents with languages, we train a modality translator~(V-L translator) to bridge the gap in the first stage. We name this training stage modality translation because the target here is to project the vision prompts into the linguistic space. To keep good compression knowledge of motion encoder and video encoder, we freeze both encoders and the LLM in this stage and the trainable part is two V-L translators only. The motion translator is a linear projection layer, and the video translator is a two-layer MLP due to the higher complexity of video data. In this modality translation stage, the training data we take is the motion captioning and video captioning data, which will be described in~\cref{sec:experiment}. 

\begin{figure}[!t]
    \centering     
    \begin{subfigure}{0.27\textwidth}
      \centering   
      \includegraphics[width=0.82\linewidth]{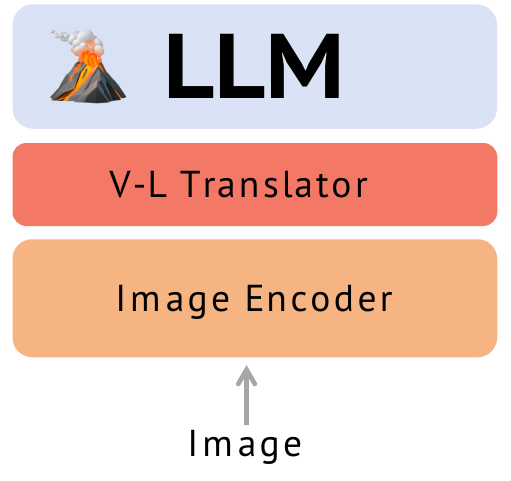}
        \caption{LLaVA Architecture.}
        \label{fig:llava}
    \end{subfigure} 
    \begin{subfigure}{0.34\textwidth}
      \centering   
      \includegraphics[width=1\linewidth]{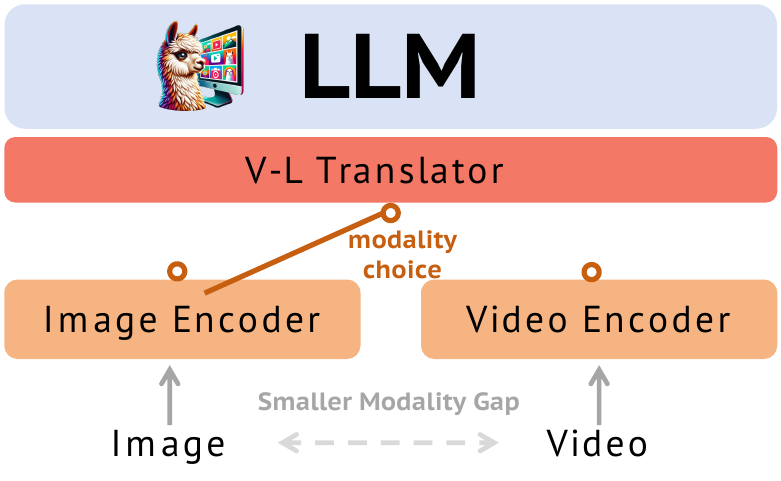}
        \caption{Video-LLaVA Architecture.}
        \label{fig:videollava}
    \end{subfigure} 
    \hspace{0.3em}
    \begin{subfigure}{0.34\textwidth}
      \centering   
      \includegraphics[width=1.03\linewidth]{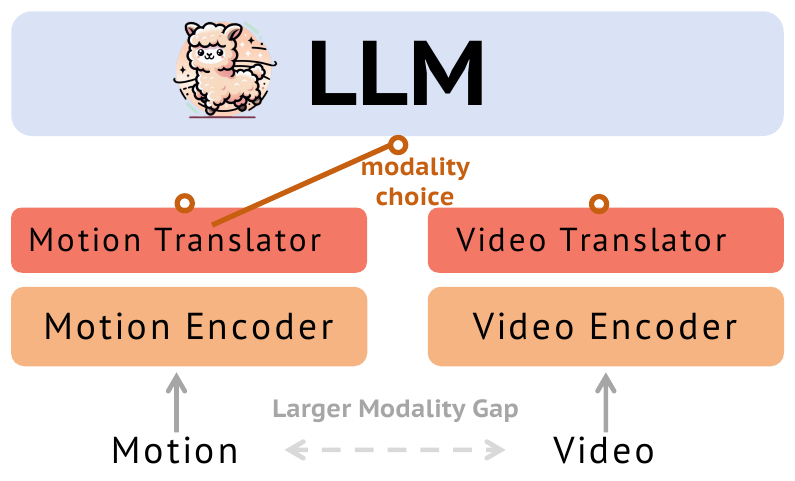}
        \caption{MotionLLM Architecture.}
        \label{fig:motionllm_compare}
    \end{subfigure} 
    \vspace{-0.5em}
    \caption{{\bf Technical comparisons  with other VLLMs.} (a) LLaVA~\cite{llava} takes the images as input only. (b) Video-LLaVA~\cite{videollava} shares a unified V-L translator for images and videos due to the small modality gap between the two modalities. (c) To bridge the larger modality gap between motion and videos, we take two separated V-L translators for better modality translations. }
    \label{fig:comparewithllavafamily}
    \vspace{-1.2em}
\end{figure}

To detail the soundness of our technical design, we compare our MotionLLM with two similar Vision LLMs~(VLLM), LLaVA~\cite{llava} and Video-LLaVA~\cite{videollava} respectively. As shown in~\cref{fig:llava}, LLaVA only takes the images as input without other external modalities. Different from LLaVA, Video-LLaVA takes images and videos as input. As can be seen in~\cref{fig:videollava}, Video-LLaVA uses different vision encoders for images and videos, respectively. As there is a small modality gap between images and videos, Video-LLaVA enjoys good performance with the shared V-L translator. However, in~\cref{fig:motionllm_compare}, motion data is a kind of structural skeleton-based data, which is quite different from pixel-level video data. This larger modality gap indicates the shared modality translator is no longer a wise choice for our task. Therefore, in MotionLLM, we take different V-L translators for motions and videos respectively. In this fashion, two modalities can enjoy better modality translation capabilities respectively.

\noindent{\textbf{Motion-video unified instruction tuning~(Stage 2).}} 
In the second stage, MotionLLM needs to respond to more diverse instructions of human inputs. Here, the visual encoders of both modalities are frozen, and the V-L translators are still trainable. Different from the training strategy in the modality translation, the LLM part $f_{\bf L}(\cdot)$ is also trainable to obtain a better understanding of visual content. To keep the original knowledge of the LLM, we train the LLM part in a parameter-efficient fine-tuning fashion~(PEFT), like LoRA~\cite{lora}. Here, with shared parameters in the LLM part, knowledge of the two modalities is interactive and shared in the linguistic space and benefits each other. Except for the careful technical design of MotionLLM, we also construct the unified instruction tuning dataset, especially motion-video-text data in pairs, which will be introduced in~\cref{sec:dataset}.


\subsection{MoVid: Human \underline{Mo}tion and \underline{Vid}eo Understanding Dataset}
\label{sec:dataset}

As discussed in~\cref{sec:intro} and~\cref{sec:arch}, we expand existing motion and video data to a unified dataset~(MoVid) for fine-grained human behavior understanding. For both motion and video parts, we construct corresponding text with captions and instructions. The details of construction and statistics are discussed as follows.

\begin{wrapfigure}{r}{0.45\textwidth}
\begin{center}
\vspace{-2.8em}
\includegraphics[width=0.8\linewidth]{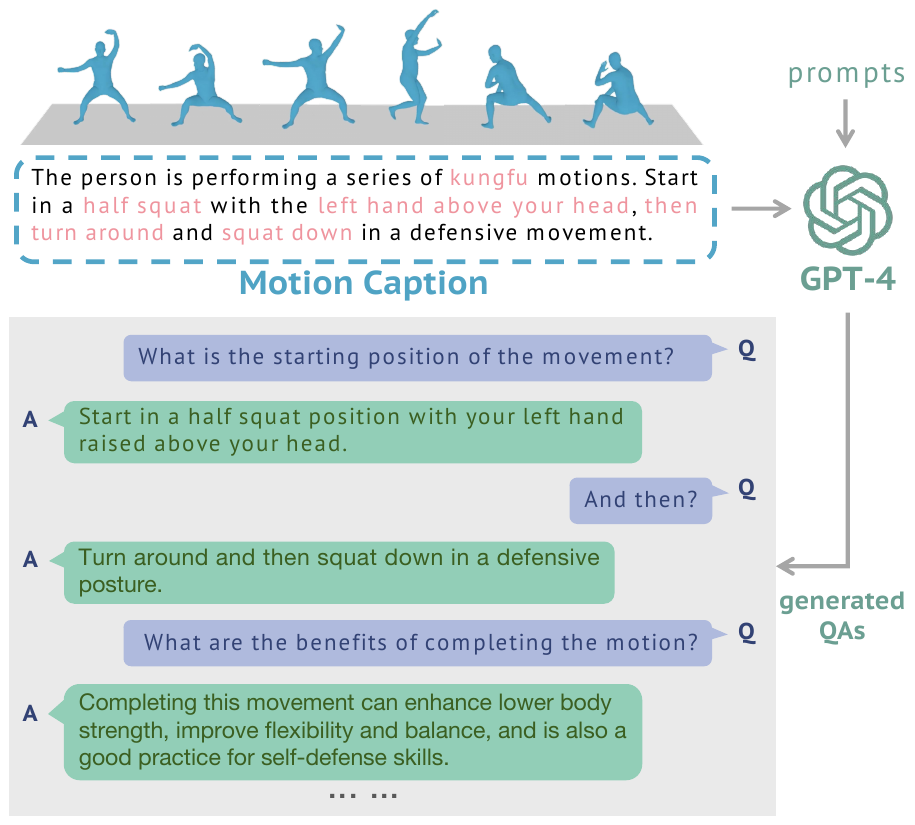}
\end{center}
\vspace{-0.8em}
\caption{{\small \bf QA of motion captioning data construction example.} We introduce GPT-4 to generate diverse QA pairs, including in-context examples~(\eg ``\texttt{And then?}''), and reasoning QAs~(\eg ``\texttt{What are the benefits of completing the motion?}'').}
\vspace{-2em}
\label{fig:gpt4h3dqa}
\end{wrapfigure} 
\noindent{\textbf{Motion-text dataset construction of MoVid.}} In the motion part of MoVid, our method mainly focuses on detailed spatial-temporal motion understanding and reasoning ability. As shown in~\cref{fig:gpt4h3dqa}, we augment the caption of HumanML3D~\cite{guo2022generating}~(\textit{a.k.a.} H3D) motion data to dialogue QAs via GPT-4~\cite{gpt4}, including 272k QA pairs in total. The generated QAs cover diverse spatial-temporal questions, in-context QAs, and reasoning data, which are used for instruction tuning. The detailed prompts and more in-context examples~\cite{zhang2023ideal} are shown in the  Appendix. Similarly to H3DQA, we also introduce the Motion-XQA instruction tuning dataset, whose caption annotation process will be detailed in the next video-text dataset construction part. The Motion-XQA comes up with 200k QA pairs in total. Different from the previous motion instruction tuning dataset~\cite{motiongpt} highly related to motion captioning, our instruction tuning dataset is more complex and diverse, including in-context examples and reasoning data.

\begin{wrapfigure}{r}{0.45\textwidth}
\begin{center}
\vspace{-2em}
\includegraphics[width=0.8\linewidth]{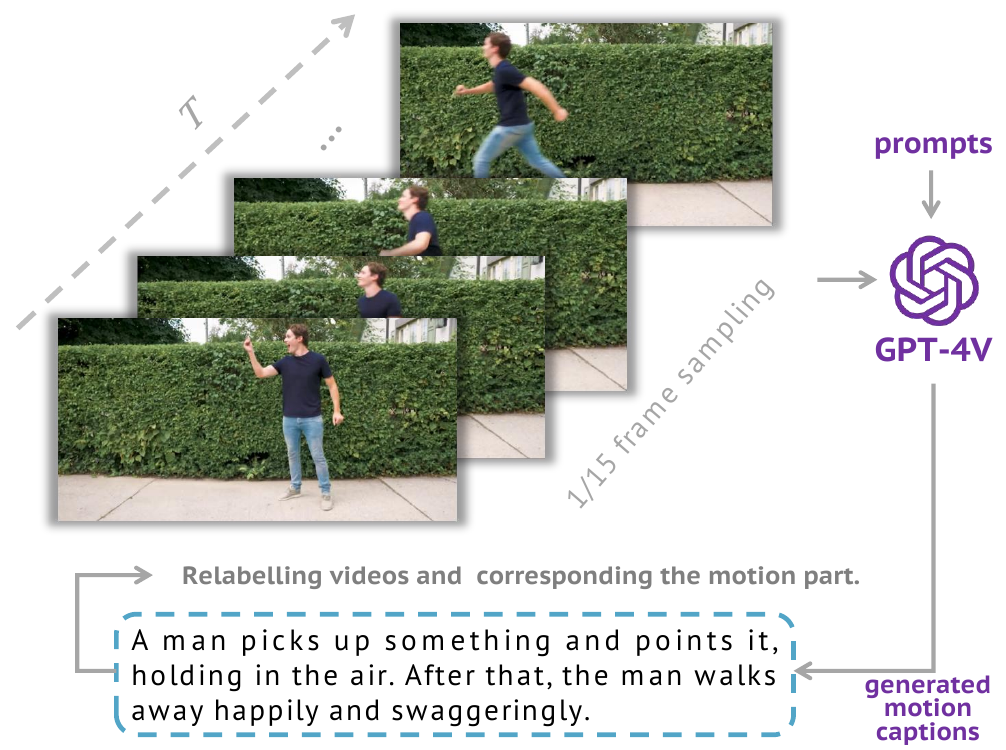}
\end{center}
\vspace{-0.8em}
\caption{{\bf GPT-4V annotation pipeline.}}
\vspace{-1.2em}
\label{fig:gpt4vann}
\end{wrapfigure} 
\noindent{\textbf{Video-text dataset construction of MoVid.}} As there are limited video-text datasets highly related to human behaviors, our main efforts mainly focus on annotating human-centric videos. Although Motion-X is with diverse motion-video pairs, its text annotation is not informative enough. To resolve this issue, as the annotation process is shown in~\cref{fig:gpt4vann}, we relabel the caption of Motion-X~\cite{motionx} via GPT-4V at first. We take the 15$\times$ down-sampling rate to extract the key frames of a video and contact them into the GPT-4V model with some carefully designed prompts~(detailed in Appendix). We check the annotated video captions and find them accurately annotated on human motions. Thanks to the \textit{pairwise} video-motion data in Motion-X~\cite{motionx}, with the well-annotated video caption data, we also relabel the caption of the motion part in Motion-X. Therefore, we can obtain 24k \textit{pairwise} motion-video data with the same textual caption, which will provide more modality alignment in the instruction tuning stage. With the obtained annotated Motion-X caption data, we generate a Motion-XQA instruction tuning dataset with multi-round QAs to empower the reasoning ability of MotionLLM. The pipeline of Motion-XQA annotation is similar to the pipeline of H3DQA (\cref{fig:gpt4h3dqa}), generated by GPT-4~\cite{gpt4}. We leave more details of Motion-XQA construction and generated examples in the Appendix. 

\begin{table*}[!t]
\begin{adjustbox}{width=1\textwidth, left}
    \begin{minipage}{0.76\linewidth}  
        \begin{tabular}{l|ccccc}
        \toprule
        Dataset          & motion & video & type & \# pairs & annotator \\
        \hline
        H3DQA            & \Checkmark      & \XSolidBrush & QA   & 272k     & GPT-4     \\
        Motion-X Caption & \Checkmark      & \Checkmark  & Caption    & 24k      & GPT-4V    \\
        Motion-XQA       & \Checkmark      & \Checkmark  & QA    & 200k     & GPT-4    \\
        \bottomrule
        \end{tabular}
        \caption{{\bf Dataset statistics.} MoVid dataset includes new caption data for Motion-X and QA pairs for H3DQA.}
        \label{tab:moviddataset}
    \end{minipage}
    
    \begin{minipage}{0.67\linewidth}
        \centering
    \tabcolsep=0.9em
        \begin{tabular}{c|c|c|c|c|c|c}
        \toprule
        & \multicolumn{5}{c}{Motion}                                     \\
        \hline
        Type & All & Body. & Seq. & Dir. & Rea. & Hall. \\
        \hline
        \# count & 700 & 205       & 171        & 140       & 148       & 36            \\
        \hline
        \hline
        & \multicolumn{5}{c}{Video}                                      \\
        \hline
        Type & All & Body. & Seq. & Dir. & Rea. & Hall. \\
        \hline
        \# count & 650 & 167       & 216        & 43        & 185       & 39           \\
        \bottomrule
        \end{tabular}
    \captionof{table}{ Benchmark statistics of Video and Motion. \label{tab:benchstatic}}
    \end{minipage}
\end{adjustbox}
\end{table*}

\noindent{\textbf{Dataset statistics.}}
We summarize the dataset we constructed for both modalities. As shown in~\cref{tab:moviddataset}, our Movid dataset includes new caption data for Motion-X and QA pairs for H3DQA and Motion-XQA. The H3DQA subset of MoVid comes up with 272k QA pairs. For Motion-X, we obtain new 24k captions of Motion-X with GPT-4V and 200k QA pairs with GPT-4. We detail more details about annotated samples in the Appendix.

\subsection{MoVid-Bench: \underline{Mo}tions and \underline{Vid}eos Understanding \underline{Bench}mark}

\begin{wrapfigure}{r}{0.3\textwidth}
\begin{center}
\vspace{-1em}
\includegraphics[width=\linewidth]{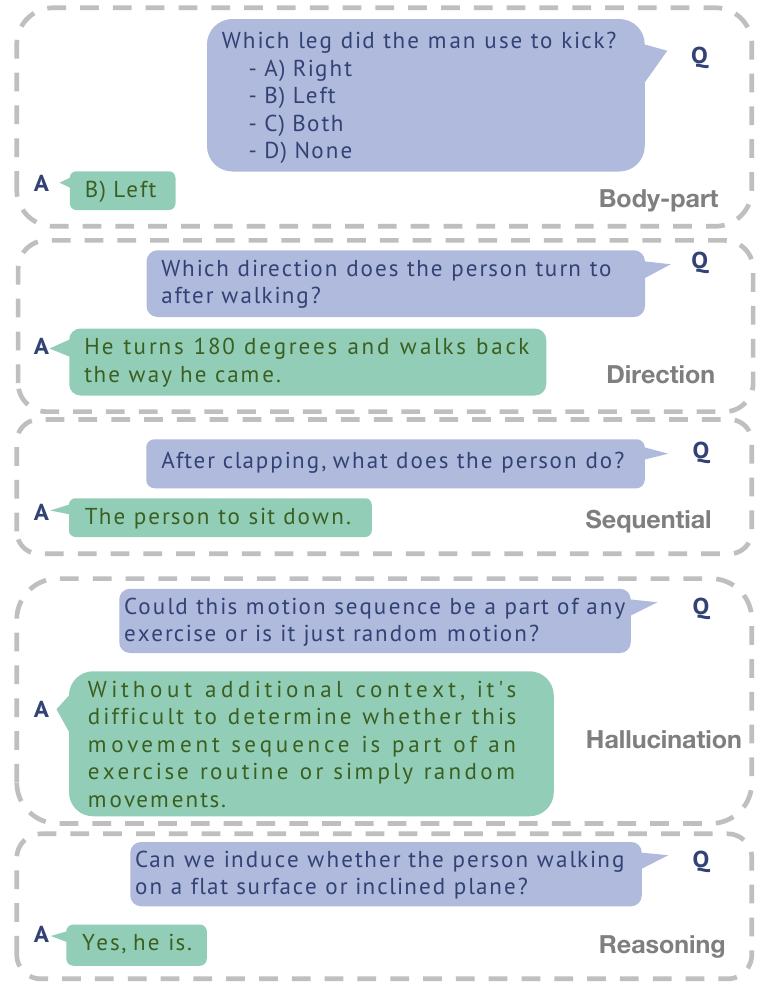}
\end{center}
\vspace{-1em}
\caption{An example of QA categorization.}
\vspace{-1em}
\label{fig:bench_vis}
\end{wrapfigure}
For a better comparison of the fine-grained human behavior understanding, we construct a benchmark for evaluating the performance, named MoVid-Bench. As shown in~\cref{tab:benchstatic}, MoVid-Bench evaluates the human behavior understanding abilities on motions and videos. Following the previous VLLM benchmark~\cite{mvbench} evaluation volumes, MoVid-Bench comes up with 1,350 data pairs, including 700 for motions and 650 for videos. For the motion part, the data is a subset of the H3DQA test set, where all QAs are carefully checked and revised by humans. Similarly, the video benchmark data is a subset of the Motion-XQA test set, where all QAs are also carefully checked and revised by humans. Besides, we design the evaluation of model performance on five aspects, including body-part motion awareness~(Body.), sequential analysis ability~(Seq.), direction awareness~(Dir.), reasoning ability~(Rea.)~\cite{sun2023survey}, and robustness against hallucination~(Hall.)~\cite{ji2023survey}, respectively. All these five aspects are categorized manually. As the movement trajectory in Motion-X videos is short, such as ``\texttt{sitting while playing guitar}'', the annotations on the direction part are limited. Besides, as the videos are given as references and the hallucination does not happen frequently, we do not evaluate this with too many examples, which is more related to the natural language processing category. We take an example in~\cref{fig:bench_vis} to show our design principle on how to categorize these five types. We leave more details and design principles in the Appendix. In~\cref{{sec:experiment}}, we will introduce our evaluation metrics on the MoVid-Bench.

\section{Experiments}
\label{sec:experiment}


\subsection{Experimental Setting}
\label{sec:expsetting}

\begin{table}
\vspace{-1em}
\centering
\begin{subtable}[t]{0.48\linewidth}
\resizebox{0.99\textwidth}{!}{
    \begin{tabular}{cc|c|c|c|c}
    \toprule
    &  \multicolumn{2}{|c|}{Stage 1}    & \multicolumn{3}{c}{Stage 2}        \\
    \hline
    \multicolumn{1}{c|}{Data} & H3D  & { \makecell{\textbf{Motion-X}\\\textbf
    {Capion}}}  & {\bf H3DQA} & {\bf Motion-XQA} & BABEL-QA \\
    \hline
    \multicolumn{1}{c|}{\# count} & 23k & 24k & 272k & 200k      & 2k    \\
    \bottomrule
    \end{tabular}   
}
    \captionsetup{font={small,stretch=0.9}}
    \caption{MoVid \textbf{motion} dataset statistics. \label{tab:statics2}}

\end{subtable}
\begin{subtable}[t]{0.48\linewidth}
    
\resizebox{0.99\textwidth}{!}{
    \begin{tabular}{c|c|c|c}
    \toprule
    & Stage 1    & \multicolumn{2}{c}{Stage 2}     \\
    \hline
    Data & Valley & {\bf Motion-XQA} & Video-ChatGPT \\
    \hline
    \# count & 702k   & 200k & 100k    \\
    \bottomrule
    \end{tabular}
    
}
\captionsetup{font={small,stretch=0.9}}
\caption{MoVid \textbf{video} dataset statistics. \label{tab:statics1}}
\end{subtable}
\vspace{-0.5em}

\caption{MoVid dataset statistics (Data in MoVid is bolded).\label{tab:staticsall}}
\vspace{-2.5em}
\end{table}
%
\textbf{Training dataset.} For motion, as shown in~\cref{tab:statics2}, we take HumanML3D~(\textit{a.k.a.} H3D) and our Motion-X Caption~(a subset of MoVid) data as our training data. In the instruction tuning stage, except for our constructed H3DQA and Motion-XQA, we additionally take 2k size BABEL-QA~\cite{babelqa} as our training data. For video, in~\cref{tab:statics1}, as we only need to learn a V-L translator in the first stage, we take the Valley~\cite{valley} video captioning dataset to train our projection layer. In the second stage, we take the Motion-XQA as a part of the training data to empower the comprehension of human behaviors. To preserve the general VQA capability, we use the Video-ChatGPT data during our instruction tuning.

\noindent{\textbf{Evaluation dataset.}} For motion understanding tasks, we evaluate motion comprehension ability on our MoVid-Bench. We also test our performance on the BABEL-QA~\cite{babelqa} test set for comparison with some expert models. For video-based tasks, we evaluate our model on three benchmarks, MVBench~\cite{mvbench}~(zero-shot), ActivityNet-QA~\cite{yu2019activitynet}~(zero-shot), and MoVid-Bench. Specifically, for the MVBench, since we do not focus on scenes and objects and for the fair evaluation of human behavior understanding, we perform a comparison of 7 human behavior-related sub-tasks, which are 1) Action Localization, 2) Action Prediction, 3) Action Sequence, 4) Egocentric Navigation, 5) Fine-grained Action, 6) Fine-grained Pose, and 7) Unexpected Action, respectively. 

\noindent{\textbf{Evaluation metrics.}}
In terms of our MoVid-Bench, following the evaluation protocol in previous research~\cite{videollava}, we utilize GPT-3.5-turbo for evaluation. Technically, the evaluation involves comparing the model answer with the ground truth answer to provide evaluation accuracy and assign a score ranging from 0 to 5. In our approach to BABEL-QA benchmark~\cite{babelqa}, following the original setting, we use prediction accuracy for evaluation. For MVBench video understanding evaluation, we answer the multi-choice questions and select the best options guided by the answer prompt ``\texttt{Best option:(}'', following~\cite{mvbench}. In this way, our model could follow the instructions well and choose the best one among the given options. In terms of ActivityNet-QA~\cite{yu2019activitynet} and our MoVid-Bench, we adopt the evaluation protocol used in~\cite{videollava, li2023videochat} by utilizing GPT-3.5-turbo, which is similar to the evaluation protocol on our MoVid-Bench.

\noindent{\textbf{Implementation details.}}
We use the lit-gpt framework~\cite{lit-gpt} and extend it to multi-modal input. We apply the pre-trained LanguageBind~\cite{languagebind} to encode video and a pre-trained VQ-VAE~\cite{t2m-gpt} encoder to encode motion data. Vicuna-7B~\cite{vicuna} is used as our base LLM model. For motion, we use a one-layer linear transformation as the motion translator to perform modality translation. For video, we use a two-layer MLP as the video translator and encode videos with 8-frame images.
When training, in the first stage, the video encoder, motion VQ-VAE encoder, and the LLM are frozen. we train the motion and video translator with a learning rate of $1\times 10^{-3}$. In the second stage, the video encoder and motion VQ-VAE encoder are still frozen; we train the video and motion translators with a learning rate of $2\times 10^{-5}$. The LLM is tuned by LoRA~\cite{lora} with a learning rate of $2\times 10^{-4}$ and the rank as $64$. In the evaluation stage, we take 8 video frames and the whole motions as the model input. We leave more training and testing details in the Appendix.

\subsection{Quantitative Results}
\vspace{-0.5em}

We show quantitative results for motion and video understanding of human behaviors on existing benchmarks and MoVid-Bench.

\begin{table}[!t]
\centering
\resizebox{\textwidth}{!}{
\begin{tabular}{l|cccccccccccc}
\toprule

\multirow{2}*{\textbf{MoVid-Bench-Motion}}& \multicolumn{2}{c|}{Body.} & \multicolumn{2}{c|}{Seq.} & \multicolumn{2}{c|}{Dir.} & \multicolumn{2}{c|}{Rea.} & \multicolumn{2}{c|}{Hall.} & \multicolumn{2}{c}{All} \\ \cline{2-13} 
& Acc.        & \multicolumn{1}{c|}{Score}    & Acc.        & \multicolumn{1}{c|}{Score}    & Acc.        & \multicolumn{1}{c|}{Score}    & Acc.        & \multicolumn{1}{c|}{Score}    & Acc.        & \multicolumn{1}{c|}{Score}    & Acc.         & Score     \\
 \hline
GT           & 100.00     & 5.00     & 100.00     & 5.00     & 100.00     & 5.00     & 100.00     & 5.00     & 100.00     & 5.00     & 100.00      & 5.00      \\
GPT-3.5~\cite{gpt3}       & 24.51      & 2.04     & 30.41      & 2.25     & 27.14      & 2.19     & 39.19      & 2.64     & \textbf{58.33}     & 3.22     & 31.33       & 2.31      \\
MotionGPT~\cite{motiongpt}    & 31.22      & \textbf{3.98}     & \textbf{42.69}      & \textbf{3.16 }    & 44.29      & 3.50     & 35.81      & 3.06     & 16.66      & 2.25     & 36.86       & 3.11      \\
\hline
MotionLLM & \textbf{50.49}      & 3.55     & 36.84      & 3.14     & \textbf{58.57}      & \textbf{3.76}     & \textbf{ 52.70}     & \textbf{3.58}     & 55.56      & \textbf{3.39}     & \textbf{49.50}\textcolor{blue}{${}_{+12.64}^{\uparrow 38\%}$}    & \textbf{3.49}\textcolor{blue}{${}_{+0.38}^{\uparrow 12\%}$}     \\

\hline
\hline      

\multirow{2}*{\textbf{MoVid-Bench-Video}}& \multicolumn{2}{c|}{Body.} & \multicolumn{2}{c|}{Seq.} & \multicolumn{2}{c|}{Dir.} & \multicolumn{2}{c|}{Rea.} & \multicolumn{2}{c|}{Hull.} & \multicolumn{2}{c}{All} \\ \cline{2-13} 
& Acc.        & \multicolumn{1}{c|}{Score}    & Acc.        & \multicolumn{1}{c|}{Score}    & Acc.        & \multicolumn{1}{c|}{Score}    & Acc.        & \multicolumn{1}{c|}{Score}    & Acc.        & \multicolumn{1}{c|}{Score}    & Acc.         & Score     \\
 \hline
GT           & 100.00     & 5.00     & 100.00     & 5.00     & 100.00     & 5.00     & 100.00     & 5.00     & 100.00     & 5.00     & 100.00      & 5.00      \\
GPT-3.5~\cite{gpt3}       &  2.40     &  1.23    &  1.39     &  1.00    &  4.65     &  1.09    &  5.41     &  1.65    & 0.00     & 0.94     &  3.03      & 1.26      \\
Video-LLAVA~\cite{videollava}    & 33.53      &   2.76   & 25.46      & 2.72    & 41.86      & 2.84     & 52.97      & 3.28     & 58.83      & 1.89     & 42.53       & 2.70      \\
\hline
MotionLLM & \textbf{34.13}      & \textbf{2.93}     & \textbf{32.87}      & \textbf{2.92}     & \textbf{44.18}      & \textbf{3.14}     & \textbf{63.20}     & \textbf{3.55}     & \textbf{70.59}      & \textbf{2.30}     & \textbf{49.00}\textcolor{blue}{${}_{+6.47}^{\uparrow 15\%}$}        & \textbf{2.97}\textcolor{blue}{${}_{+0.27}^{\uparrow 10\%}$}     \\

\bottomrule

\end{tabular}
}
\vspace{0.2em}
\caption{{\bf Comparison on the MoVid-Bench.} The top table is for motion and the bottom table is for video. The larger the accuracy and score, the better the result. }
\label{tab:motionmain}
\end{table}

\begin{table}[!t]
\centering
\scriptsize
\resizebox{\textwidth}{!}{
\begin{tabular}{l|c|c|ccc|ccc}
\toprule \specialrule{0em}{0pt}{0pt}
Model & Pred. type & Overall $\uparrow$ & Action $\uparrow$ & Direction $\uparrow$ & Body Part $\uparrow$ & Before $\uparrow$ & After $\uparrow$ & Other $\uparrow$ \\
\hline
2s-AGCN-M~\cite{2sagcn} & cls. & 0.355 & 0.384 & 0.352 & 0.228 & 0.331 & 0.264 & 0.295 \\
2s-AGCN-R~\cite{2sagcn} & cls.   & 0.357 & 0.396 & 0.352 & 0.194 & 0.337 & 0.301 & 0.285 \\
MotionCLIP-M~\cite{motionclip} & cls.  & 0.430 & 0.485 & 0.361 & \textbf{0.272 }& 0.372 & 0.321 & 0.404 \\
MotionCLIP-R~\cite{motionclip} & cls.  & 0.420 & 0.489 & 0.310 & 0.250 & 0.398 & 0.314 & 0.387 \\
\hline
MotionLLM & gen. & 0.372 & 0.396 & \textbf{0.417} & 0.154 & 0.329 & 0.353 & 0.338 \\
MotionLLM* & gen. & \textbf{0.436} & \textbf{0.517} & 0.354 & 0.154 & \textbf{0.427} & \textbf{0.368} & \textbf{0.529} \\
\specialrule{0em}{0pt}{0pt}
\bottomrule
\end{tabular}
}
\vspace{0.2em}
\caption{{\bf Comparison of different methods on BABEL-QA test set.} The ``*'' denotes finally fine-tuned on BABEL-QA. ``Pred. type'' denotes the prediction type, including closed set classification (cls.) and open vocabulary generation (gen.). ``-M'' and ``-R'' denote MLP and RNN, respectively. MotionLLM shows comparable performance with \textbf{close-set regression expert models}.}
\label{tab:babelqa}
\end{table}

\noindent{\textbf{Evaluation motion understanding capability on MoVid-Bench.}} We compare MotionLLM with baselines on MoVid-Bench (motion part) on five aspects: body-part awareness, sequentially, direction analysis, reasoning ability, and hallucination, respectively. The evaluation follows previous LLM evaluation metrics~\cite{videollava,li2023videochat,jin2023chat} on accuracy and scores. We compare our method with text-only GPT-3.5 answer results and MotionGPT results. As shown in~\cref{tab:motionmain}, our model performs the best results with baselines on overall accuracy and scores. As the GPT-3.5 baseline cannot be compatible with motions, it cannot understand human motion accurately. Specifically, MotionGPT shows limited reasoning and robustness against hallucination. MotionGPT is trained on the HumanML3D dataset only, and the instruction tuning dataset mainly focuses on the motion caption task, like ``\texttt{Describe the motion represented by <Motion> in plain English.}'' or ``\texttt{What does the <Motion> communicate? Please describe it in language.}''. This instruction-tuning dataset makes it hard to follow complex instructions, like reasoning or fine-grained spatial-temporal understanding. MotionLLM relieves these issues, benefiting from our carefully designed instruction tuning dataset.

\noindent{\textbf{Evaluation on BABEL-QA.}} 
We additionally show the spatial-temporal capacity of MotionLLM on BABEL-QA, which includes diverse spatial-temporal questions. Following~\cite{babelqa}, we compare MotionLLM with several baselines. 
1) \textbf{2s-AGCN}, an end-to-end approach using 2s-GCN to extract motion features and predict the answer with an MLP~(-M) or an RNN~(-R). 2) \textbf{MotionCLIP}, a transformer-based method used for extracting motion features and predicting the answer with an MLP~(-M) or an RNN~(-R). Note that these baselines answer the questions of BABEL-QA in a closed vocabulary set. We take two stages MotionLLM model for comparison. Here, we set the prediction accuracy as the evaluation metric. As the evaluation is the exact string matching, we set up a baseline MotionLLM* fine-tuned on BABEL-QA. From~\cref{tab:babelqa}, although our method is an open vocabulary multi-modality language generation model, MotionLLM still enjoys comparable performance with expert models. As the exact string matching is not reasonable for our final model, we also compare our final MotoinLLM with baselines via GPT evaluation, similar to the setting introduced in~\cref{sec:expsetting}. As shown in~\cref{tab:babelqa}, MotionLLM also shows comparable accuracy with baselines. The performance drop in the first stage mainly comes from the modality and task compromise.

\begin{table}[!t]
\centering
\scriptsize
\resizebox{\textwidth}{!}{
\begin{tabular}{l|c|c|ccccccc|c}
\toprule
Model                         & \multicolumn{1}{c|}{LLM}   & \multicolumn{1}{c|}{Frames}     & \multicolumn{1}{c}{AL}      & \multicolumn{1}{c}{AP}      & \multicolumn{1}{c}{AS}      & \multicolumn{1}{c}{EN}      & \multicolumn{1}{c}{FA}      & \multicolumn{1}{c}{FP}      & \multicolumn{1}{c|}{UA}   & {Avg.}   \\ \hline
Otter-V~\cite{otter}                      & \llama-7B                   & 16                                                  & 23.5                        & 23.0                        & 23.0                        & 23.5                        & 27.0                        & 22.0                        & 29.5     & 24.5                   \\
mPLUG-Owl-V~\cite{mplug}                 & \llama-7B                   & 16                                                  & 23.0                        & 28.0                        & 22.0                        & 26.0                        & 29.0                        & 24.0                        & 29.0     & 25.8                   \\
VideoChatGPT~\cite{maaz2023video}              & Vicuna-7B                  & 100                                                 & 20.0                        & 26.0                        & 23.5                        & 29.5                        & 22.5                        & 29.0                        & 26.5      & 25.2                  \\
VideoLLaMA~\cite{zhang2023video}                  & Vicuna-7B                  & 16                                                  & 22.5                        & 25.5                        & 27.5                        & \textbf{30.0}               & 29.0                        & \textbf{32.5}               & 39.0  & 29.4                      \\
VideoChat~\cite{li2023videochat}                  & Vicuna-7B                  & 16                                                  & 27.0                        & 26.5                        & \textbf{33.5}               & 23.5                        & \textbf{33.5}               & 26.5           & \textbf{40.5} & 30.1              \\
Video-LLaVA~\cite{videollava}                & Vicuna-7B                  & 8                                                 & 22.5                        & 25.5                        & 29.5                        & 29.0                        & 24.5                        & 28.5                        & 24.5     & 26.3                     \\
\hline
MotionLLM                & Vicuna-7B                  & 8                                         & \textbf{33.0}               & \textbf{29.5}                        & 32.5                 & 29.0             & 31.5                        & 28.5        & 37.5      & \textbf{31.6}\textcolor{blue}{$_{+1.5}^{\uparrow 5\%}$}                   \\ \bottomrule
\end{tabular}
}
\vspace{0.3em}
\caption{{\bf Comparison with different video-based LLMs on MV-Bench.} MotionLLM outperforms baselines on overall average metric. }
\label{tab:mvbench}
\end{table}

\noindent{\textbf{Evaluation video understanding results on MoVid-Bench.}}
\cref{tab:motionmain} shows the evaluation results on MoVid-Bench~(video part) across five key areas defined in the MoVid-Bench construction part. This evaluation adheres to the established metrics used in previous large language model (LLM) evaluation~\cite{videollava,li2023videochat,jin2023chat}. As can be seen in~\cref{tab:motionmain}, without video grounds, it is hard for GPT-3.5 to answer the questions well. Our MotionLLM significantly outshines the previously leading video model, Video-LLaVA~\cite{videollava}, in each evaluated aspect. Specifically, MotionLLM achieves a $\uparrow$15\% improvement in average accuracy and a $\uparrow$10\% enhancement in average score over Video-LLaVA. The performance of Video-LLaVA is weaker in sequentiality, reasoning, and hallucination tasks due to its lack of joint training with motion data. Conversely, MotionLLM makes reasonable use of motion data alongside a carefully designed instruction tuning dataset, leading to enhanced performance. We provide more discussion on more video-based LLMs on our MoVid-Bench in the Appendix.

\noindent{\textbf{Zero-shot video multi-choice QA on MVBench.}}
As shown in~\cref{tab:mvbench}, we conduct a \textbf{zero-shot} evaluation of video question-answering capabilities on MVbench~\cite{mvbench} with 7 different human motion-related sub-tasks. Despite processing merely eight frames per video, our MotionLLM model surpasses existing video-based LLM baselines in terms of average accuracy. Notably, MotionLLM achieves a $\uparrow$5\% higher average accuracy than its closest competitor, VideoChat~\cite{li2023videochat}. In the areas of Action Localization and Action Prediction, our model demonstrates superior performance over all competitors, highlighting its exceptional ability to understand temporal dynamics. Remarkably, MotionLLM leads VideoChat by $\uparrow$22\% in Action Localization, underscoring its effectiveness in compensating for the limitations of a video encoder that only processes eight frames. Moreover, the performance on other sub-tasks is on par with that of other baselines, indicating that our model maintains the spatial relationships and reasoning capabilities.

\begin{wraptable}{r}{0.5\textwidth}
\centering
\vspace{-1.3em}
\resizebox{0.37\textwidth}{!}{
\begin{tabular}{l|cc}
\toprule
\textbf{Methods} & \textbf{Acc.} & \textbf{Score} \\ \hline
FrozenBiLM & 24.7 & - \\ 
VideoChat & - & 2.2 \\ 
LLaMA-Adapter & 34.2 & 2.7 \\ 
Video-LLaMA & 12.4 & 1.1 \\ 
Video-ChatGPT & 35.2 & 2.7 \\ 
Video-LLaVA & 45.3 & 3.3 \\ 
Video-chat2 & 49.1 & 3.3 \\ 
\hline 
MotionLLM & \textbf{53.3}\textcolor{blue}{$_{+4.2}^{\uparrow9\%}$} & \textbf{3.52}\textcolor{blue}{$_{+0.22}^{\uparrow7\%}$} \\ \bottomrule
\end{tabular}
}
\vspace{-0.6em}
\captionsetup{font={small,stretch=0.8}}
\caption{{\bf Results on ActivityNet-QA.}}
\vspace{-2em}
\label{table:activitynet}
\end{wraptable}
\noindent{\textbf{Zero-shot open-vocabulary video QA on ActivityNet-QA.}}
To better evaluate our model on human behaviors for long videos, we conduct \textbf{zero-shot} evaluations on ActivityNet-QA, in~\cref{table:activitynet}. Note that our model is not trained with any ActivityNet data. Our MotionLLM surpasses the leading model by $\uparrow$9\% on accuracy and $\uparrow$7\% on score metrics, showing good sequential understanding and generalizable human behavior comprehension of the video content. This indicates the promising application of MotionLLM in the real world.

\subsection{Qualitative Results}

\begin{figure}[!t]
    \centering
    \includegraphics[width=\textwidth]{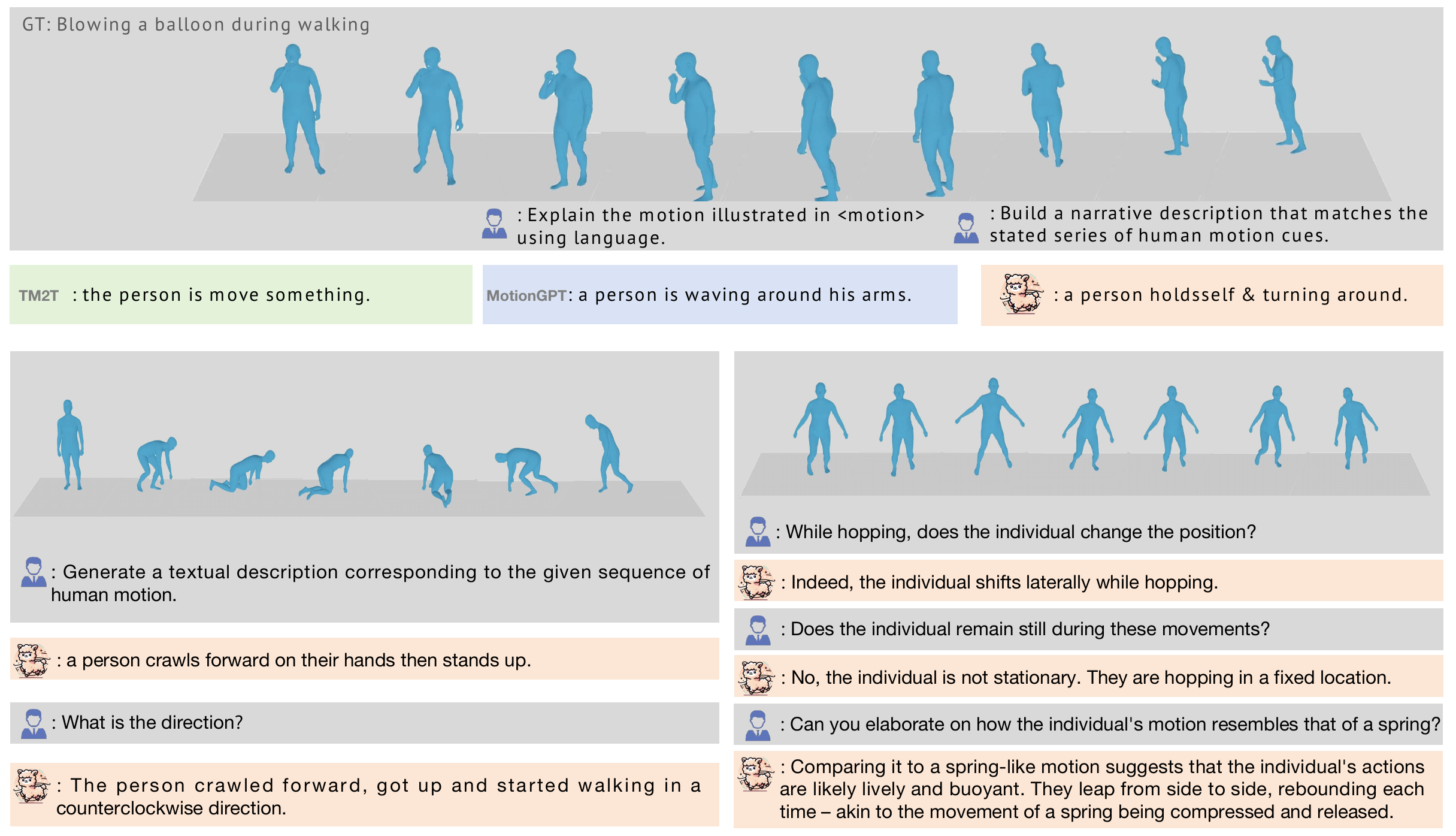}

    \caption{{\bf Examples of motions comprehension.} The results demonstrate the proficiency of MotionLLM in captioning, spatial-temporal understanding, and reasoning. Comparison with TM2T~\cite{tm2t} and MotionGPT~\cite{motiongpt} underscores the effectiveness of MotionLLM in handling unseen motions. 
    }
    \label{fig:motion_main_vis}
\end{figure}

\noindent{\textbf{Qualitative results on motion-based comprehension.}} For the understanding of MotionLLM on the motion modality, we provide more visualization results in~\cref{fig:motion_main_vis}. For the first comparison with TM2T~\cite{tm2t} and MotionGPT~\cite{motiongpt}, we choose the MotionLLM model checkpoint trained without Motion-X data for comparison. The comparison is conducted on the IDEA-400 subset of Motion-X~\cite{motionx} in a zero-shot test setting. As can be seen in~\cref{fig:motion_main_vis}, MotionLLM enjoys more generalization ability on some in-the-wild scenarios and has the potential to work as an automatic text annotator for motion data. We further take some in-context examples to show the spatial awareness of MotionLLM. For the final case, MotionLLM not only understands the motion patterns but also enjoys good reasoning capabilities for motion comprehension.

\begin{figure}[!t]
    \centering
    \includegraphics[width=\textwidth]{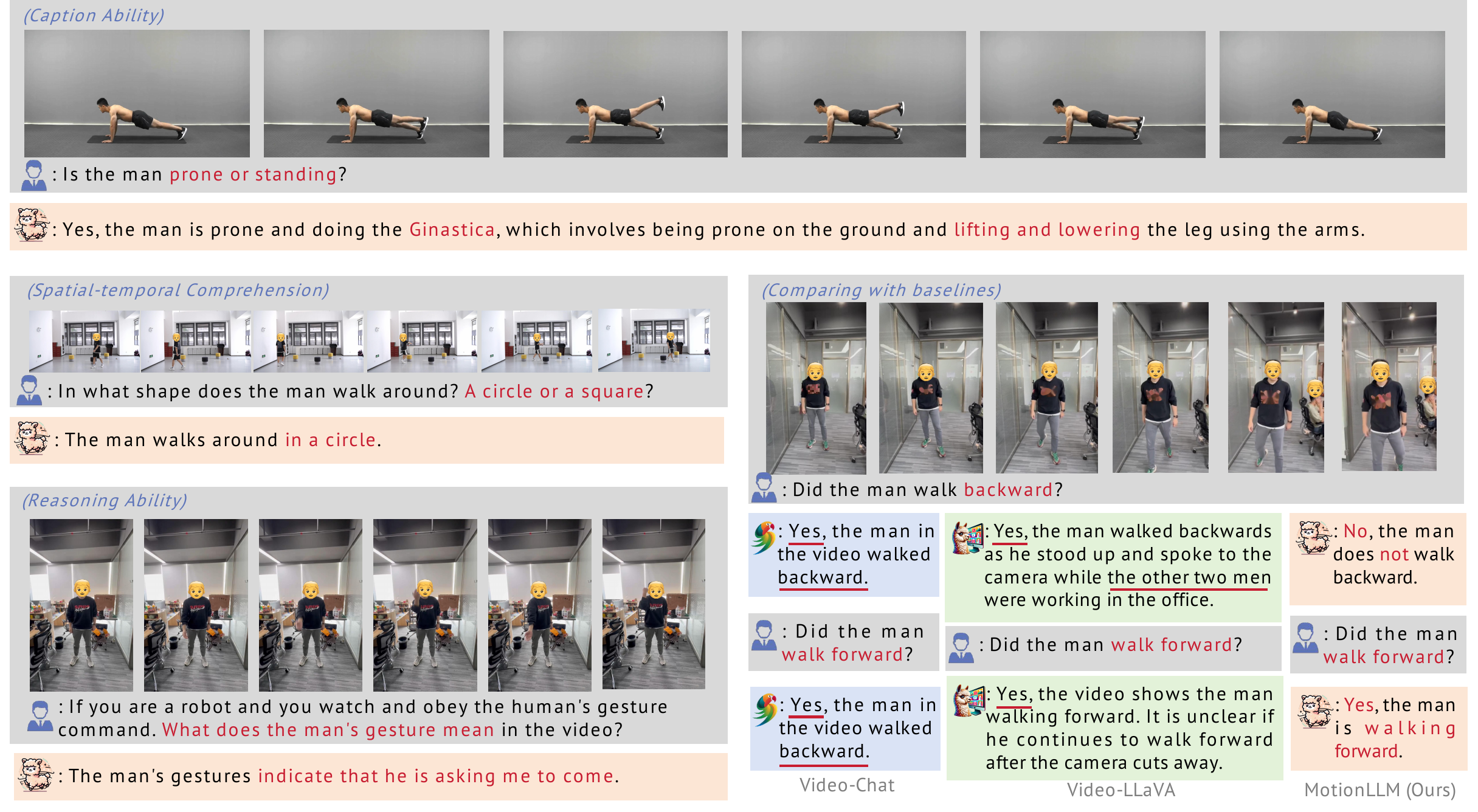}
    \caption{{\bf Videos comprehension of models.} The results show good performance of MotionLLM on captioning, spatial-temporal comprehension, and reasoning. The comparison with Video-Chat~\cite{li2023videochat} and Video-LLaVA~\cite{videollava} shows good sequentiality and direction comprehension of MotionLLM.}
    \label{fig:video_main_vis}
\end{figure}

\noindent{\textbf{Qualitative results on video-based comprehension.}} To analyze our language outputs on understanding human behaviors from videos, we take some representative examples to explore the human behavior understanding capability of MotionLLM in~\cref{fig:video_main_vis}. Expect the basic captioning capacity. Our MotionLLM also enjoys good spatial-temporal comprehension, such as ``\texttt{walks around a circle}''. Moreover, thanks to the basic reasoning ability of LLM, MotionLLM can also induce the intention of human behaviors, such as inducing the ``\texttt{indicate that he is asking me to come}'' intention from the becking motion, showing the potential of application in embodied intelligence scenarios. We additionally perform a comparison with Video-Chat~\cite{li2023videochat} and Video-LLaVA~\cite{videollava} on the temporal understanding ability. Although Video-Chat can answer the first question correctly, its second answer is contradictory, failing to obtain good in-context learning capability. Besides, Video-LLaVA fails in the first question and always answers the ``\texttt{Yes}''. Different from these methods, MotionLLM enjoys better in-context learning capability and temporal comprehension than baselines.

\vspace{-1em}
\subsection{Ablation Study}

Here, we conduct ablation studies on different modality modeling strategies and show results of instruction tuning using unpaired data from H3DQA, BabelQA, and Video-ChatGPT instruction data and paired data Motion-XQA discussed above. Note that the ``paired'' data statement here denotes the Motion-XQA subset of MoVid, including motion-video-text triple pairs. This dataset design aims to make the understanding of video and motion benefit from each other. The performance is tested on our benchmark MoVid-Bench.

\noindent{\textbf{Ablation on motion understanding.}}
As seen in the top part of the~\cref{tab:ablation}, the usage of video data helps to improve the motion understanding overall, especially on the body description, reasoning abilities, and hallucination reduction. With the help of videos, the overall performance improved by 28.6\% on average accuracy. We attribute this to the fact that video provides more reference information, such as human-environment interaction information for motion modality. When instruction tuning with unpaired video-motion data, the abilities of all five aspects are improved, indicating the advantages of the joint training strategy with videos. Moreover, based on this, the usage of our paired data MotionX-QA boosts the performance further in most aspects, except for the sequential perception ability. We argue this is due to the limitation of the video encoder compression capacity, which can only encode 8 frames, losing too much information. Therefore, when training with more videos using our MotionX-QA, the motion branch will compromise to this limitation and be affected. Recent progress~\cite{sora} in video compression might be a promising fashion to reveal this.

\noindent{\textbf{Ablation on video understanding.}}
As can be seen in the bottom part of the~\cref{tab:ablation}, incorporating unpaired motion datasets such as H3DQA and BABEL-QA has proven to enhance the sequential perception capabilities for the video branch significantly. The improvement of other capabilities is limited. The modest effect is mainly due to the limited amount of H3DQA and BABEL-QA data. Upon conducting additional instruction tuning with our paired dataset, Motion-XQA, we observed holistic enhancements in all five aspects, culminating in a notable 17\% improvement in overall accuracy. It indicates the effectiveness of joint training with paired motion-video data, enabling the model to more adeptly utilize motion cues and enhance the integration by transferring information across the different modalities.

\begin{table}[!t]
\centering
\resizebox{\textwidth}{!}{
\begin{tabular}{cccc|cccccccccccc}
\toprule

\multicolumn{4}{c|}{\multirow{1}*{MoVid-Bench-Motion}}& \multicolumn{2}{c|}{Body.} & \multicolumn{2}{c|}{Seq.} & \multicolumn{2}{c|}{Dir.} & \multicolumn{2}{c|}{Rea.} & \multicolumn{2}{c|}{Hall.} & \multicolumn{2}{c}{All} \\ \cline{0-2} \cline{2-16} 

Motion & Video & Unpair & Pair & Acc.        & \multicolumn{1}{c|}{Score}    & Acc.        & \multicolumn{1}{c|}{Score}    & Acc.        & \multicolumn{1}{c|}{Score}    & Acc.        & \multicolumn{1}{c|}{Score}    & Acc.        & \multicolumn{1}{c|}{Score}    & Acc.         & Score     \\
 \hline
 \Checkmark & \XSolidBrush & - & - & 35.29   &  2.77 & 39.18 & 2.90 & 53.57 & 3.25 & 34.45 & 2.77 & 11.11 & 2.00 & 38.48 & 2.86\\
\Checkmark       & \Checkmark      & \Checkmark     & \XSolidBrush      & 47.55 & 3.54 & \textbf{46.20} & \textbf{3.26} & 46.43 & 3.49 & \textbf{53.38} & \textbf{3.63} & 44.44 & 3.08 & 48.07 & \textbf{3.50} \\
\hline
\Checkmark       & \Checkmark &  \Checkmark & \Checkmark & \textbf{50.49}      & \textbf{3.55}     & 36.84      & 3.14     & \textbf{58.57}      & \textbf{3.76}     & { 52.70}     & {3.58}     & \textbf{55.56}      & \textbf{3.39}     & \textbf{49.50}\textcolor{blue}{${}_{+11.02}^{\uparrow 29\%}$}        & {3.49}\textcolor{blue}{${}_{+0.63}^{\uparrow 8\%}$}     \\

\hline
\hline      

\multicolumn{4}{c|}{\multirow{1}*{MoVid-Bench-Video}}& \multicolumn{2}{c|}{Body.} & \multicolumn{2}{c|}{Seq.} & \multicolumn{2}{c|}{Dir.} & \multicolumn{2}{c|}{Rea.} & \multicolumn{2}{c|}{Hall.} & \multicolumn{2}{c}{All} \\ \cline{0-2} \cline{2-16} 

Motion & Video & Unpair & Pair & Acc.        & \multicolumn{1}{c|}{Score}    & Acc.        & \multicolumn{1}{c|}{Score}    & Acc.        & \multicolumn{1}{c|}{Score}    & Acc.        & \multicolumn{1}{c|}{Score}    & Acc.        & \multicolumn{1}{c|}{Score}    & Acc.         & Score     \\
 \hline
 \XSolidBrush & \Checkmark & - & - & 33.53     & 2.76     & 25.46     & 2.72     & 41.86     & 2.84     & 52.97     & 3.28     & 58.83     & 1.89     & 42.53      & 2.70     \\
\Checkmark       & \Checkmark      & \Checkmark     & \XSolidBrush      & 31.74     & 2.80      & 28.70     & 2.69      & 32.56     & 2.78     & 49.73     & 3.21       & 64.71 & 2.29 & 41.49  & 2.75 \\
\Checkmark       & \Checkmark &  \Checkmark & \Checkmark & \textbf{34.13}      & \textbf{2.93}     & \textbf{32.87}      & \textbf{2.92  }   & \textbf{44.18}      & \textbf{3.14}     & \textbf{63.20}     & \textbf{3.55 }    & \textbf{70.59}      & \textbf{2.30}     & \textbf{48.94}\textcolor{blue}{${}_{+6.41}^{\uparrow 15\%}$}        & \textbf{2.97}\textcolor{blue}{${}_{+0.27}^{\uparrow 10\%}$}  \\

\bottomrule

\end{tabular}
}
\vspace{0.2em}
\caption{{\bf Ablation studies for modeling different datasets and modalities.} The top table is for motion and the bottom table is for video. Unpair refers to using unpaired instruction datasets, including H3DQA, BabelQA, Video-ChatGPT instruction datasets, while Pair means using Motion-XQA to do instruction tuning.}
\label{tab:ablation}
\end{table}

\section{Conclusion and Discussion}
\label{sec:conclusion}
\noindent{\textbf{Conclusion.}} In this work, we have presented MotionLLM, a unified framework for human behavior understanding, focusing on human motion and video modalities. MotionLLM introduces an LLM-based framework to bridge the gap among motions, videos, and languages. To empower good spatial-temporal understanding and reasoning capabilities, we constructed a MoVid dataset to include diverse question-answer pairs of motions and videos on spatial-temporal understanding and reasoning. We also developed a MoVid-Bench to evaluate the understanding capability of models on human behaviors. Experiments show the effectiveness of both our methods and datasets on fine-grained human behavior understanding. 

\noindent{\textbf{Limitation and Impact Statement.}} This work suffers from the limited capacity of the video encoder. Future work may consider improving the capacity of video encoders. MotionLLM is promising to serve as an AI assistant in many scenarios, like a fitness coach for social goods, especially for the visually impaired community. For the negative impact, the development of LLMs might raise the possibility of negative use of our model, such as negative content on social media.

\section*{Acknowledgement}

The author team would like to deliver many thanks to many people. Qing Jiang helps a lot with some parts of manual annotation on MoVid Bench and resolves some ethics issues of MotionLLM. Jingcheng Hu provided some technical suggestions for efficient training. Shilong Liu and Bojia Zi provided some significant technical suggestions on LLM tuning. Jiale Liu, Wenhao Yang, and Chenlai Qian provided some significant suggestions for us to polish the paper. Hongyang Li helped us a lot with the figure design. Yiren Pang provided GPT API keys when our keys were temporarily out of quota.

\clearpage


\newpage

\appendix

\begin{center}
{\Large Appendix for \\ \logo \ MotionLLM: Understanding Human Behaviors \\from Human Motions and Videos}
\end{center}

\section{More Comparisons on the MoVid-Bench}

We compare MotionLLM with more video-based LLMs in this section. Due to the page limits, we leave more compassion results on MoVid-Bench in the appendix. All these results are the average of three evaluations. 

Overall, MotionLLM obtains state-of-the-art results on our MoVid-Bench video part indicating the effectiveness of our model and architecture design. In addition, VideoChat2~\cite{mvbench} could achieve the best on the body description part, while our MotionLLM could achieve the best on other parts. Due to our joint training with motion data, our model could get substantial improvement in direction perception and reasoning aspects. 

\begin{table}[!h]
\resizebox{\textwidth}{!}{
\begin{tabular}{l|cccccccccccc}
\toprule
& \multicolumn{2}{c|}{Body.} & \multicolumn{2}{c|}{Seq.} & \multicolumn{2}{c|}{Dir.} & \multicolumn{2}{c|}{Rea.} & \multicolumn{2}{c|}{Hall.} & \multicolumn{2}{c}{All} \\ \cline{2-13} 
& Acc.        & \multicolumn{1}{c|}{Score}    & Acc.        & \multicolumn{1}{c|}{Score}    & Acc.        & \multicolumn{1}{c|}{Score}    & Acc.        & \multicolumn{1}{c|}{Score}    & Acc.        & \multicolumn{1}{c|}{Score}    & Acc.         & Score     \\
 \hline
GT           & 100.00     & 5.00     & 100.00     & 5.00     & 100.00     & 5.00     & 100.00     & 5.00     & 100.00     & 5.00     & 100.00      & 5.00      \\
GPT-3.5~\cite{gpt3}       &  2.40     &  1.23    &  1.39     &  1.00    &  4.65     &  1.09    &  5.41     &  1.65    & 0.00     & 0.94     &  3.03      & 1.26      \\
Video-LLAVA~\cite{videollava}    & 33.53      &   2.76   & 25.46      & 2.72    & 41.86      & 2.84     & 52.97      & 3.28     & 58.83      & 1.89     & 42.53       & 2.70      \\
Video-LLaMA~\cite{zhang2023video}    & 32.90      &   2.81   & 28.20      & 2.81    & 41.87      & 2.95     & 59.46      & 3.42    & 53.95      & 1.89     & 43.28       & 2.77      \\
VideoChat~\cite{li2023videochat}    & 41.21      &   2.93   & 28.21      & 2.81    & 32.73      & 2.78    & 46.34      & 3.15     & 62.13      & 2.21     & 42.12       & 2.79      \\
VideoChat2~\cite{mvbench}    & \textbf{43.11}      &   \textbf{3.11}   & 31.60      & 2.91    & 34.88      & 3.05     & 48.65      & 3.22     & 64.71      & 2.12     & 44.59       & 2.88      \\

\hline
MotionLLM & 34.13      & 2.93     & \textbf{32.87}      & \textbf{2.92}     & \textbf{44.18}      & \textbf{3.14}     & \textbf{63.20}     & \textbf{3.55}     & \textbf{70.59}      & \textbf{2.30}     & \textbf{49.00}       & \textbf{2.97}    \\ 
\bottomrule
\end{tabular}
}
\vspace{0.3em}
\caption{{\bf More comparisons on the MoVid-Bench (video part).} The larger the accuracy and score, the better the result. }
\label{tab:video_movid_bench}
\end{table}

\label{sec:benchres_appen}

\clearpage

\section{Technical Details}
\label{sec:implementation}

\subsection{Implementation Details}
\textbf{Model training.} During the first stage of modality translation, we trained both the motion and video translators on the NVIDIA Tesla A100-80GB GPU, utilizing the AdamW optimizer with a weight decay of 0.01. The motion translator underwent training for 40k iterations, whereas the video translator was trained for 70k iterations to accommodate the varying amounts of data in their respective datasets. In the second stage, for the motion-video unified instruction tuning, we trained the LoRA and the two translators on the NVIDIA A100-80GB GPU with a batch size of 2 on each GPU for a single epoch, requiring 96 hours. The training still employed the AdamW optimizer with a weight decay of 0.01. For training on unpaired datasets, we sampled only one modality per batch, ensuring that all samples within a batch belonged to the same modality. Conversely, for the paired datasets training (specifically for MotionX-QA), each batch contained one motion instruction QA and one video instruction QA.

\noindent{\textbf{Model inference.}} All testing and inference tasks were performed on a single NVIDIA A100-80GB GPU. We repeated all tests three times to calculate the mean results.

\subsection{Evaluation Details}

Our MoVid-Bench evaluation extends the evaluation protocol of previous multi-modality LLMs evaluations~\cite{videollava}. The evaluation of GPT will return a dictionary of predictions and a score. The details of the evaluation prompt are shown in~\cref{tab:gpt_eval}.

\begin{table*}[!h]
\centering
\begin{minipage}{0.99\columnwidth}\vspace{0mm}    \centering
    \begin{tcolorbox} 
        \raggedright
        \small
\textcolor{darkgreen}{\textbf{Input: } \texttt{question}, \texttt{answer}, \texttt{prediction} \\}
\textcolor{darkgreen}{\textbf{LLM evaluation prompts:}\\}
You are an intelligent chatbot designed for evaluating the correctness of generative outputs for question-answer pairs. \\
Your task is to compare the predicted answer with the correct answer and determine if they match meaningfully. Here's how you can accomplish the task: \\
------\\
\#\#INSTRUCTIONS: \\
- Focus on the meaningful match between the predicted answer and the correct answer.\\
- Consider synonyms or paraphrases as valid matches.\\
- Evaluate the correctness of the prediction compared to the answer.\\
Please evaluate the following video-based question-answer pair:\\
Question: \{\textcolor{darkgreen}{\texttt{question}}\}\\
Correct Answer: \{\textcolor{darkgreen}{\texttt{answer}}\}\\
Predicted Answer: \{\textcolor{darkgreen}{\texttt{prediction}}\}\\
Provide your evaluation only as a yes/no and score where the score is an integer value between 0 and 5, with 5 indicating the highest meaningful match. \\
Please generate the response in the form of a Python dictionary string with keys 'pred' and 'score', where value of 'pred' is a string of 'yes' or 'no' and value of 'score' is in INTEGER, not STRING.\\
DO NOT PROVIDE ANY OTHER OUTPUT TEXT OR EXPLANATION. Only provide the Python dictionary string. \\
For example, your response should look like this: \{'pred': 'yes', 'score': 4.8\}.
    \end{tcolorbox}
    \vspace{-0.5em}
    \caption{{\bf GPT evaluation prompts.}}
    \label{tab:gpt_eval}
\vspace{-5mm}
\end{minipage}
\end{table*}

\newpage

\section{Dataset Construction of MoVid}

\subsection{Constructing H3DQA and Motion-XQA Dataset}

\noindent{\textbf{Prompts and Response.}} We detail a template prompt we use to generate the H3DQA, and Motion-XQA instruction tuning dataset. As shown in~\cref{tab:h3dqatemplate}, our prompt comes up with our requirements and some in-context examples. The Motion-X QA instruction tuning dataset construction is similar to the annotation process of H3DQA.

\noindent{\textbf{Construction pipeline and dataset post-processing.}} To obtain the QAs according to the prompt in~\cref{tab:h3dqatemplate}, we concatenate the prompt and the motion caption as input for GPT-4. For post-processing the obtained response string, we process the string to question-answer pairs via a language parser. We ignore these very few cases for responses not aligned with the prompt command. The whole annotation process is detailed in~\cref{algorithm:intructionsetcons}.

\subsection{Motion-X Recaption Using GPT-4V}

As the textual annotation of the existing Motion-X dataset is too coarse, we relabel the caption of the Motion-X dataset. As videos in the Motion-X dataset are pairwise with motions, the relabelled caption can also used as the motion captions. As shown in~\cref{tab:gpt4vannpprompt}, our prompt comes up with our requirements, where the \texttt{general\_description} is the vanilla coarse caption of Motion-X data. Besides, the images from the video with a 15$\times$ down-sampling rate are also fed into the GPT-4V. The annotation process is shown in~\cref{algorithm:intructionsgpt4v}.

\begin{algorithm}[!h]
    \renewcommand{\algorithmicrequire}{\textbf{Input:}}
	\renewcommand{\algorithmicensure}{\textbf{Output:}}
	\caption{GPT-4 Instruction Tuning Dataset Construction.}
    \label{algorithm:intructionsetcons}
    \begin{algorithmic}
        \REQUIRE \texttt{\textcolor{teal}{prompts}}, \texttt{\textcolor{teal}{motion\_captions}} .
        \ENSURE QA pairs \verb|answer|.
        \STATE \texttt{\textcolor{teal}{input\_string}} = \texttt{\textcolor{teal}{prompts}} + \texttt{\textcolor{teal}{motion\_captions}}
        \STATE    completion = openai.ChatCompletion.create(
        \STATE    \ \ \ \ model=``gpt-4'',
        \STATE    \ \ \ \ messages=[
        \STATE    \ \ \ \ \ \ \ \ \{``role'': ``user'', ``content'': \texttt{ \textcolor{teal}{input\_string}}\},
        \STATE    \ \ \ \  ]
        \STATE         )
        \STATE   \verb|answer_string| = completion.choices[0].message["content"]
        \STATE   \verb|answer| = \verb|parser|(\verb|answer_string|)
        \STATE   {\textbf{Return}}  \verb|answer|
    \end{algorithmic} 
\end{algorithm}

\vspace{-1em}
\begin{algorithm}[H]
    \renewcommand{\algorithmicrequire}{\textbf{Input:}}
	\renewcommand{\algorithmicensure}{\textbf{Output:}}
	\caption{GPT-4 Recaption on the Motion-X}
    \label{algorithm:intructionsgpt4v}
    \begin{algorithmic}
        \REQUIRE \texttt{\textcolor{teal}{prompts}}, \texttt{\textcolor{teal}{images}} .
        \ENSURE \verb|caption|.
        \STATE \verb|content| = []
        \STATE \verb|content|.\verb|append|({``type'': ``text'', ``text': \texttt{\textcolor{teal}{prompts}}})
        \STATE \verb|content|.\verb|append|({``type'': ``image\_url'', ``text'':  ``image\_url'': \{``url'': f``data:image/jpeg;base64,\{\verb|resize_image|(\verb|down_sample|(\texttt{\textcolor{teal}{images}}), width, height)}''\}\})
        \STATE    \verb|messages| = [\{
        \STATE    \ \ \ \ ``role'': ``user'',
        \STATE    \ \ \ \ ``content'': content
        \STATE    \}]
        \STATE   \verb|payload| = \{
        \STATE    \ \ \ \ ``model'': ``gpt-4-vision-preview'',
        \STATE    \ \ \ \ ``messages'': \verb|messages|,
        \STATE    \}
        \STATE   \verb|response| = \verb|requests|.\verb|post|(f``{api\_base}/chat/completions'', headers=headers, json=\verb|payload|, proxies=proxies)
        \STATE   \verb|caption| = \verb|response|.json()[``choices''][0][``message''][``content'']
        \STATE   {\textbf{Return}}  \verb|caption|
    \end{algorithmic} 
\end{algorithm}

\begin{table*}[!h]
\vspace{-1em}
\centering
\begin{minipage}{0.99\columnwidth}\vspace{0mm}    \centering
    \begin{tcolorbox} 
        \raggedright
        \small
\textcolor{darkgreen}{\textbf{Prompt example:}\\}
This is multiple descriptions of ONE motion sequence, each line is a description, and the last two numbers are the starting time. Please construct several QA pairs based on this. If they are all 0.0, it represents the entire sequence, otherwise, it corresponds to the corresponding starting time.
\\! Note that the starting time of each row only corresponds to the description of this row, and has nothing to do with the descriptions of other rows. Avoid ambiguity.
\\! Note that you can only see the action, not the descriptions in advance. !Therefore, you can NOT ask or answer something like “the first/third description”
\\! Note that Don't say anything like ‘sure’ or ‘here is xxx’, just return QA to me directly.
\\Don’t rigidly imitate the template either
\\
**HERE IS AN EXAMPLE:**\\ 
$[$GIVEN DESCRIPTION$]$
\\
a person walks toward the front, turns to the right, bounces into a squat, and places both arms in front of chest before placing them on the knees.\#0.0\#5.0
\\person walks up and squats slightly to pose a position\#0.0\#0.0
\\he moved forward then stretched his body, moving his hand and touching his knees with hands and now he turned to the left side.\#0.0\#0.0
\\a person walks, turns slightly to the right, squats, puts hand on both knees while squatting, and then squats again.\#0.0\#8.0 \\
$[$System Output$]$: \\
**Q** What is the motivation behind the person performing these motions?
\\
**A** Without additional context, it is not possible to determine the exact motivation. The sequence could be part of an exercise routine, a movie, or a specific functional movement.
\\
**Q** True or False: The person touches their knees with their hands immediately after walking.
\\
**A** False. He stretches his body before touching his knees.
\\
**Q** What does the man do after walking forward?
\\
**A** The guy turns to the right.
\\
**Q** And then?
\\
**A** He bounces into a squat.
\\
**Q** How many times does the man do this?
\\
**A** twice.
\\
**Q** Can you describe the initial movement of the person in this sequence and the timing associated with it? Is there any repetition in the sequence, and how does this affect the overall flow?
\\
**A** Initially, the person begins the sequence by walking forward. The timing for this initial movement is marked as 0.0, indicating that it is the beginning of the sequence. The answer to the second question is YES. There is a repetition of the squatting movement within the sequence. This repetition emphasizes the squat as a core element of the sequence, reinforcing its importance. It also adds a rhythmic aspect to the movement, creating a pattern that could be aesthetically pleasing or functionally significant in terms of exercise or dance routines.
\\
**Q** Could you elaborate on the arm movements of the person during the squat and their significance in the sequence?
\\
**A** While in the squat position, the person executes a detailed arm movement. Initially, they place both arms in front of their chest, which could be interpreted as a preparatory or balancing action. Subsequently, they shift their arms to rest on their knees. This transition in arm positioning is crucial as it demonstrates coordination and control, integrating the upper body into the sequence, which was initially focused on lower body movements.
    \end{tcolorbox}
    \vspace{-0.8em}
    \caption{{\bf An example of instruction tuning dataset (H3DQA, and Motion-XQA) construction.} Our prompt comes up with our requirements and some in-context examples.}
    \label{tab:h3dqatemplate}
\end{minipage}
\end{table*}

\begin{table*}[!h]
\vspace{-0.5em}
\centering
\begin{minipage}{0.99\columnwidth}\vspace{0mm}    \centering
    \begin{tcolorbox} 
        \raggedright
        \small
\textcolor{darkgreen}{\textbf{Prompt example:}\\}
You'll be shown different frames, which are uniformly sampled from one human motion video. \\
You will also be given a general description of the video. In this video, the general description is: \{general\_description\}. \\
Please tell me what the person in the video doing and its detailed description. \\
Notice: you need to combine all the frames of the same motion video and the given general description as if you are actually seeing the video. \\
And give a description of the temporal joint movements and spatial body part movements. 
Your description should focus more on the atomic action of different body parts. It is necessary to describe the temporal sequence of different body parts, 
such as: A man took a step forward, and at the same time, thrust forward, then returned to the original position. 
Tell me what the person doing in less than 50 words, please don't be redundant.
    \end{tcolorbox}
    \vspace{-0.8em}
    \caption{{\bf An example of Motion-X re-caption annotation process.} Our prompt comes up with our requirements.}
    \label{tab:gpt4vannpprompt}
\vspace{-5mm}
\end{minipage}
\end{table*}

\clearpage

\subsection{MoVid Dataset Overall Annotation Process Summary}

\begin{figure}[!h]
    \centering
    \includegraphics[width=\textwidth]{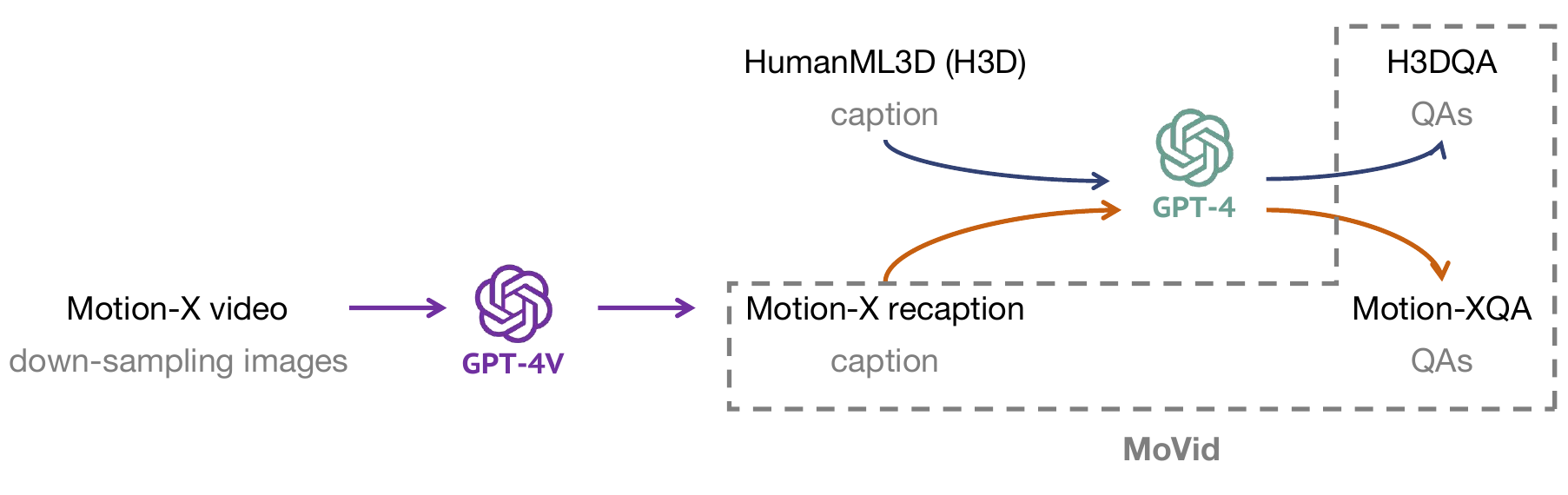}
    \vspace{-2em}
    \caption{{\bf The whole annotation process of MoVid.} 
    }
    \label{fig:appendix_ann_all}
\end{figure}

We summarize the whole annotation process of the MoVid dataset in~\cref{fig:appendix_ann_all}. As stated in the main paper, we regenerate Motion-X captions with GPT-4V at first and augment them as QAs via GPT-4. The H3DQA is augmented based on HumanML3D captions via GPT-4, too. 

\subsection{MoVid Dataset Samples}

We provide some examples of the generated MoVid dataset, including H3DQA, Motion-X caption, and Motion-XQA. 

\begin{figure}[!t]
    \centering
    \includegraphics[width=0.75\textwidth]{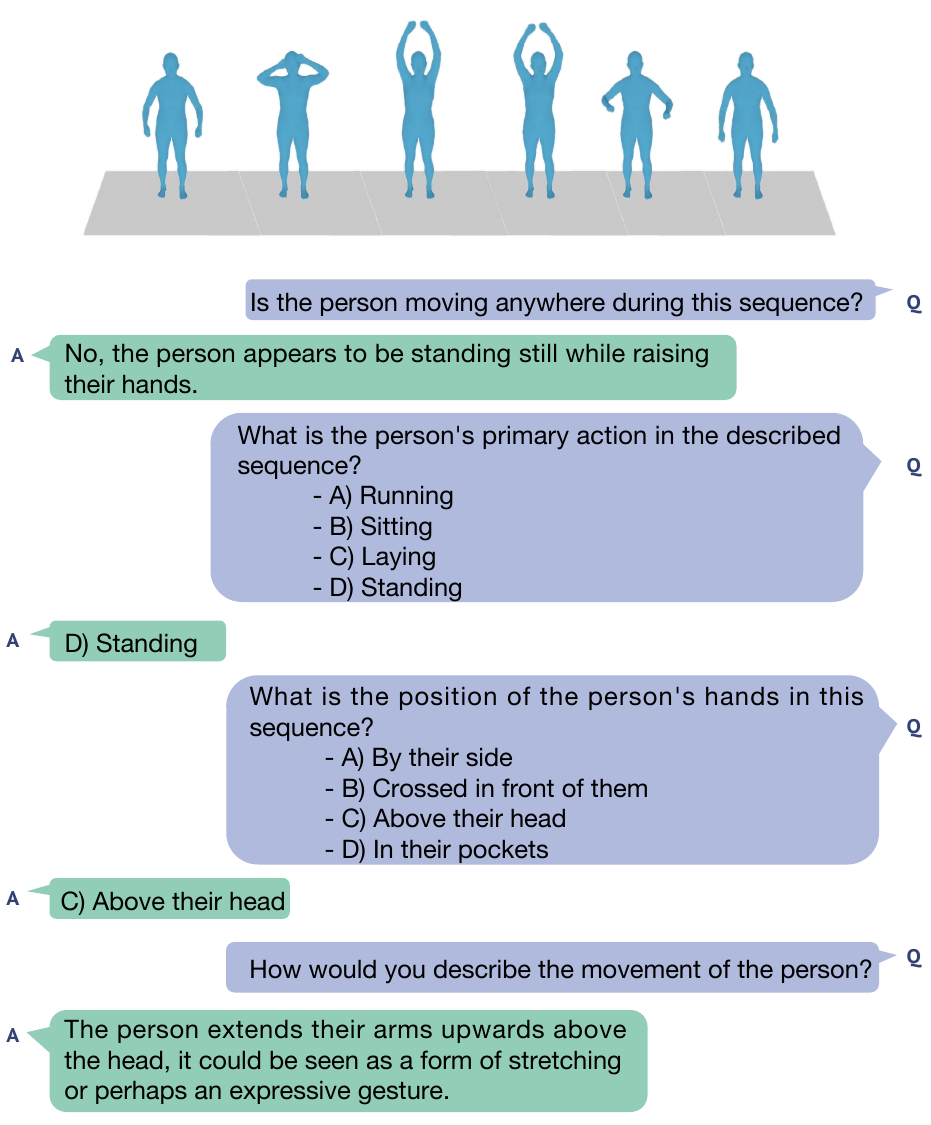}
    \caption{{\bf H3DQA sample (example 1).} 
    }
    \label{fig:h3dqa_sample_1}
\end{figure}

\begin{figure}[!t]
    \centering
    \includegraphics[width=0.75\textwidth]{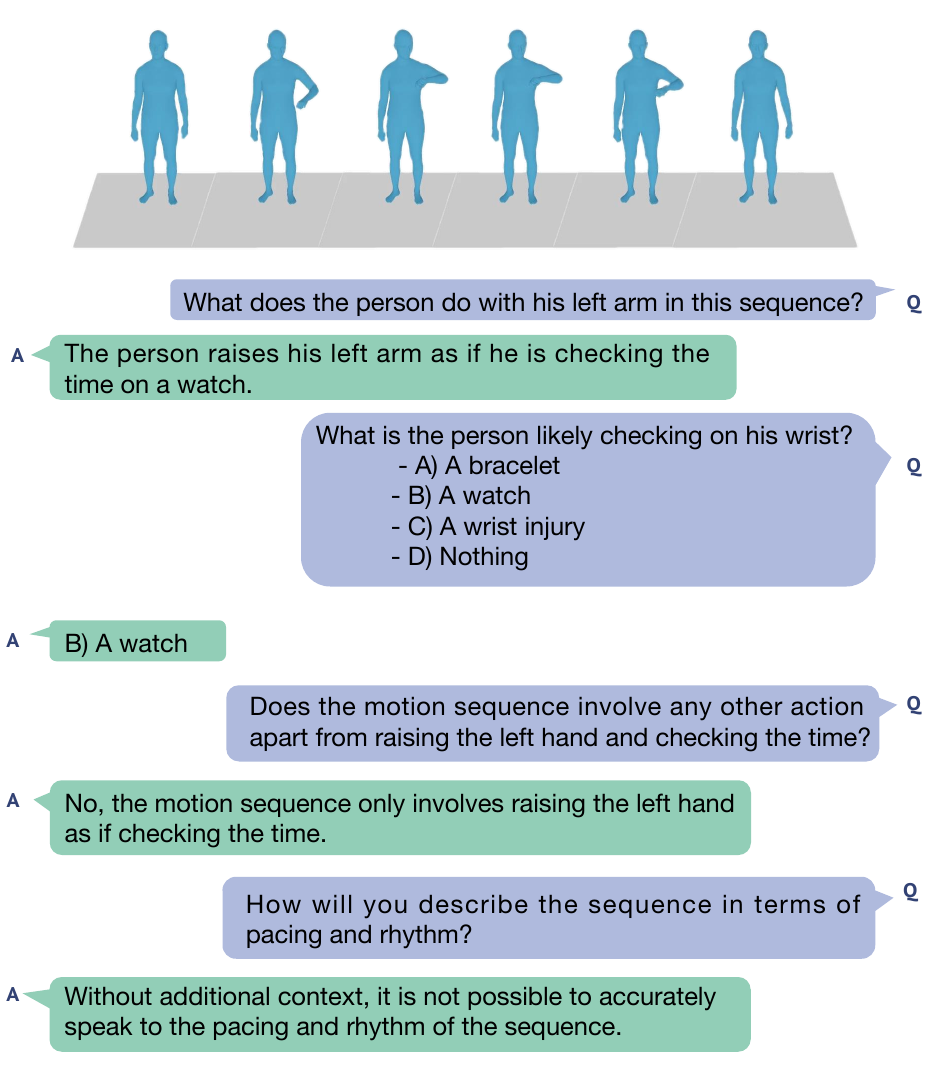}
    \caption{{\bf H3DQA sample (example 2).} 
    }
    \label{fig:h3dqa_sample_2}
\end{figure}

\noindent{\textbf{H3DQA data samples.}} Here, we present some samples of H3DQA generated by GPT-4~\cite{gpt4}. The H3DQA includes multi-round question-answer pairs, related to captioning, in-context QAs, and reasoning data. Please refer to~\cref{fig:h3dqa_sample_1} and~\cref{fig:h3dqa_sample_2}.

\clearpage

\begin{figure}[!h]
    \centering
    \includegraphics[width=\textwidth]{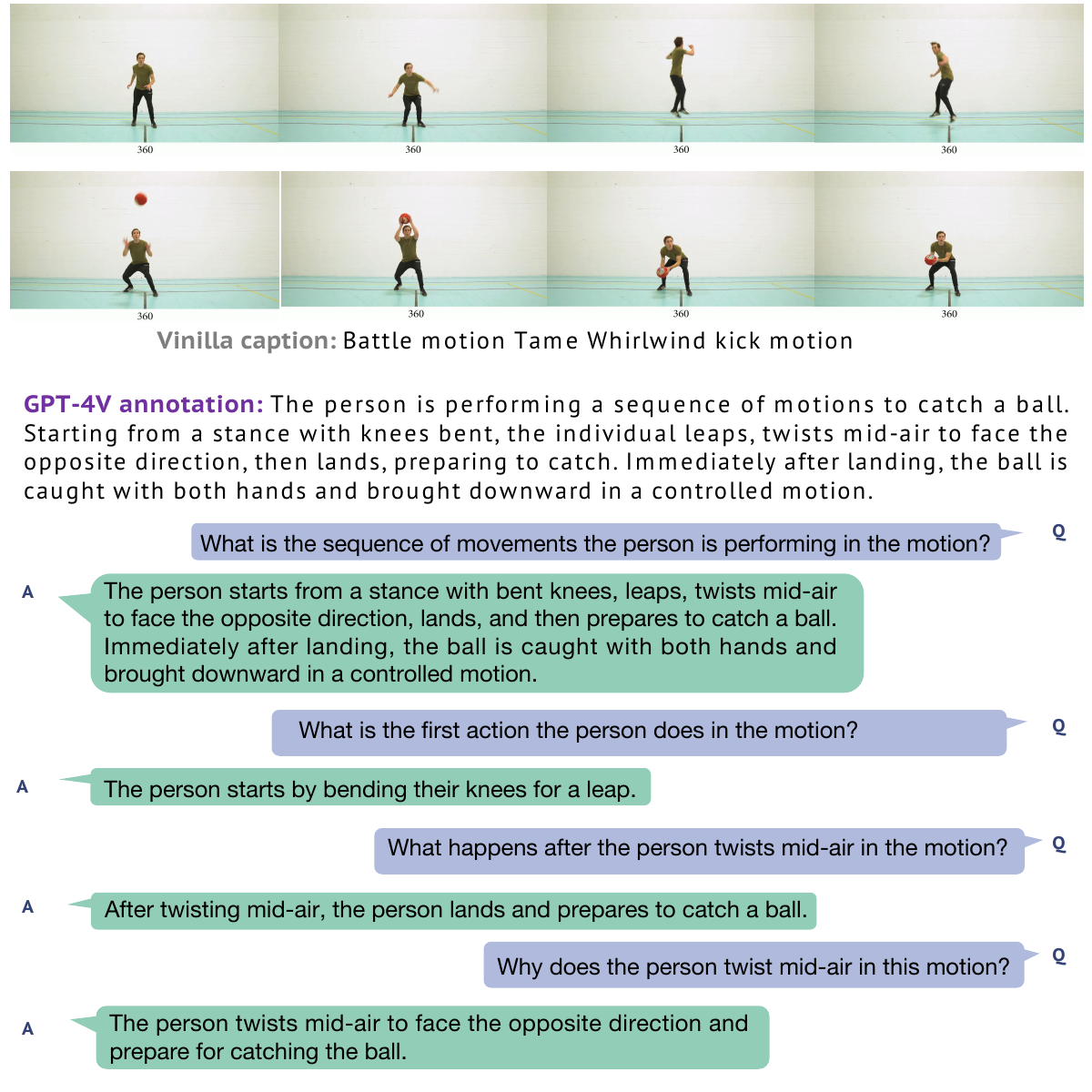}
    \caption{{\bf Motion-X reception and Motion-XQA (sample 1).} 
    }
    \label{fig:motionxcap_sample_1}
\end{figure}

\begin{figure}[!t]
    \centering
    \includegraphics[width=\textwidth]{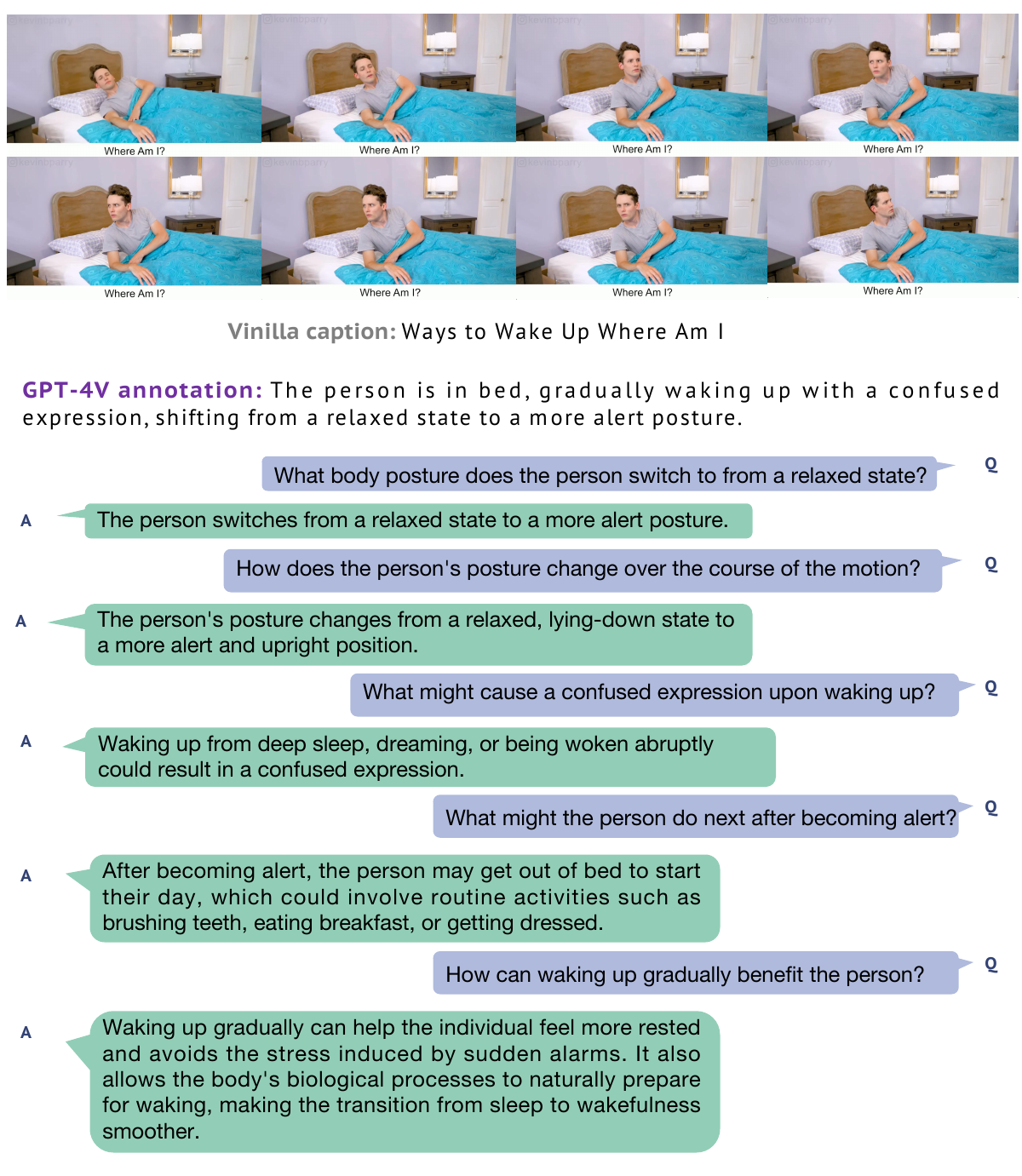}
    \caption{{\bf Motion-X reception and Motion-XQA (sample 2).} 
    }
    \label{fig:motionxcap_sample_2}
\end{figure}

\begin{figure}[!t]
    \centering
    \includegraphics[width=\textwidth]{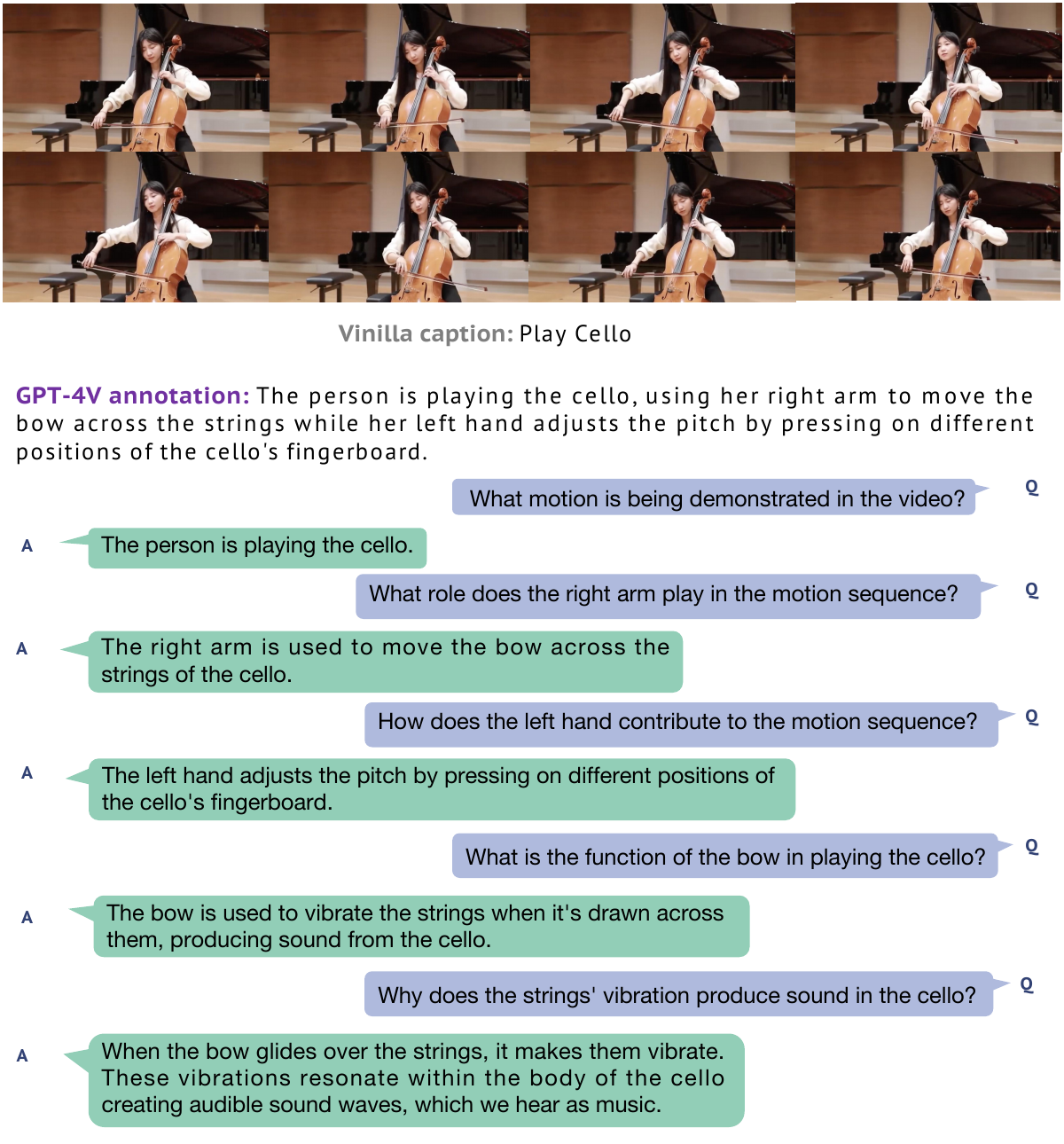}
    \caption{{\bf Motion-X reception and Motion-XQA (sample 3).} 
    }
    \label{fig:motionxcap_sample_3}
\end{figure}

\noindent{\textbf{Motion-X recaption and Motion-XQA data samples.}} We present some samples of Motion-X new caption generated by GPT-4V~\cite{gpt4} and Motion-XQA data generated by GPT-4~\cite{gpt4}. The Motion-X new caption data shows more high-quality motion caption data than previous annotations. The Motion-XQA includes multi-round question-answer pairs, related to captioning, in-context QAs, and reasoning data. Please refer to~\cref{fig:motionxcap_sample_1}, \cref{fig:motionxcap_sample_2}, and~\cref{fig:motionxcap_sample_3}.

\clearpage

\subsection{Details and Design Principles of MoVid-Bench}

We detail our design principles of MoVid-Bench. As stated in the main paper, our MoVid-Bench benchmark mainly focuses on evaluating body-part motion awareness~(Body.), sequential analysis ability~(Seq.), direction awareness~(Dir.), reasoning ability~(Rea.)~\cite{sun2023survey}, and robustness against hallucination~(Hall.)~\cite{ji2023survey}, respectively. The body-part motion awareness and direction awareness aim to evaluate the spatial understanding ability of human motions. The sequential analysis focuses on the temporal comprehension ability of the model. The reasoning ability is a basic evaluation of the LLMs and analysis of the intelligence of models. The hallucination is a revelation of LLM-based models, which mainly rely on the capability of based LLMs. 

\begin{figure}[!h]
    \centering
    \includegraphics[width=0.7\textwidth]{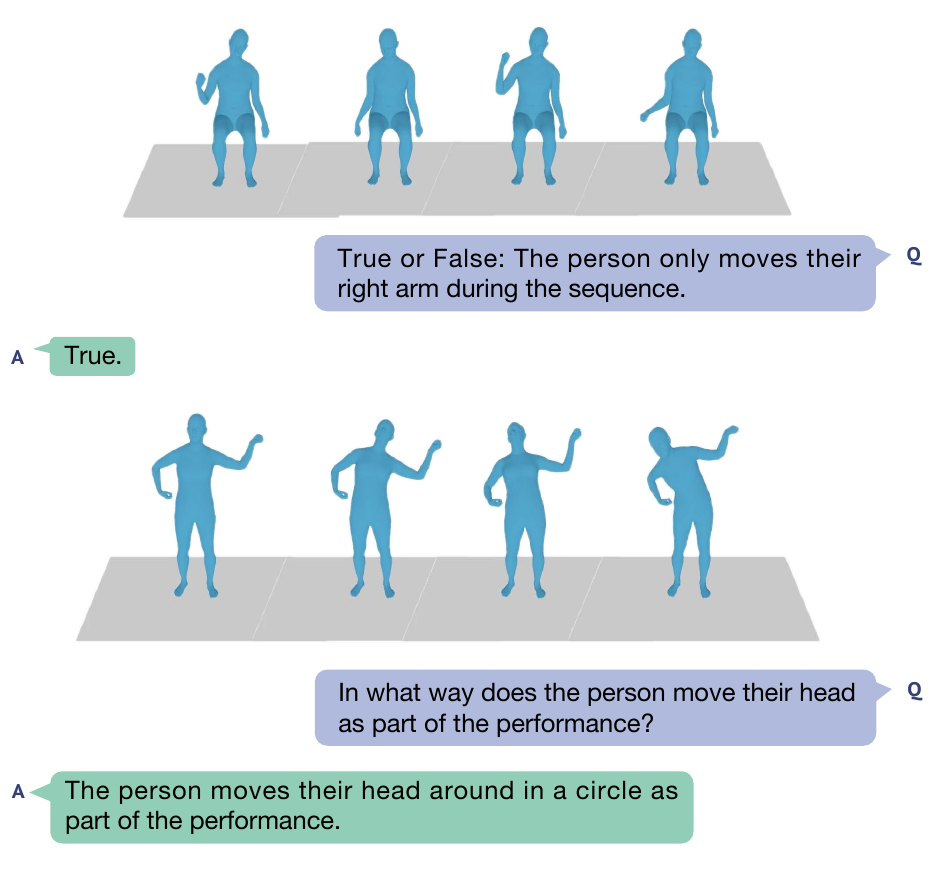}
    \caption{{\bf MoVid-Bench samples (body-part motion awareness).} 
    }
    \label{fig:bench_example_body}
\end{figure}

\begin{figure}[!t]
    \centering
    \includegraphics[width=0.7\textwidth]{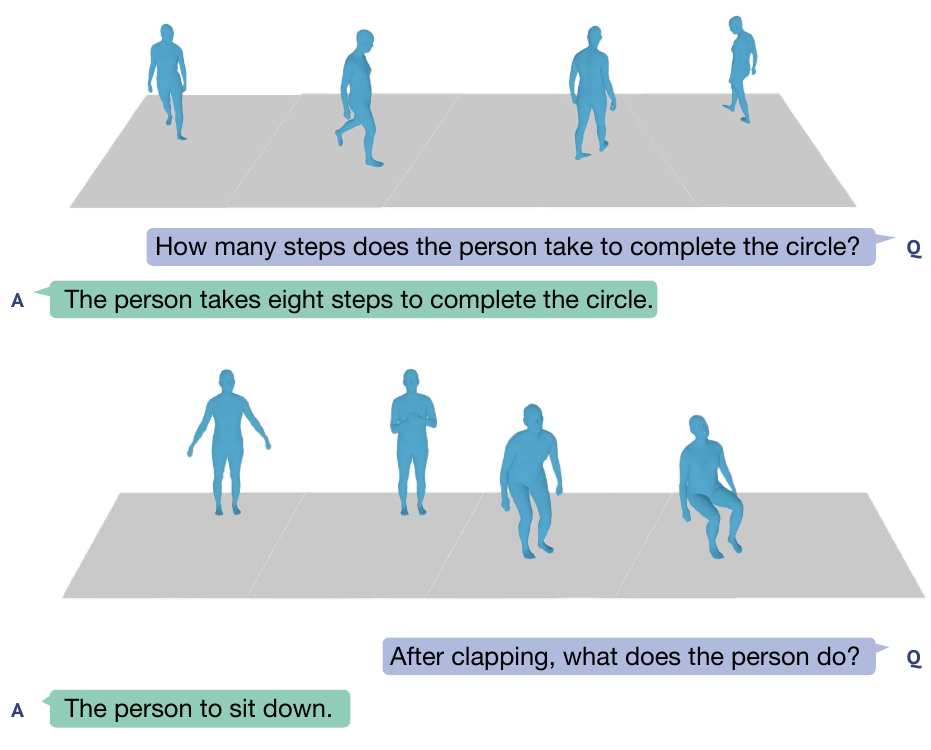}
    \caption{{\bf MoVid-Bench samples (sequential analysis ability).} 
    }
    \label{fig:bench_example_seq}
\end{figure}

\begin{figure}[!t]
    \centering
    \includegraphics[width=0.7\textwidth]{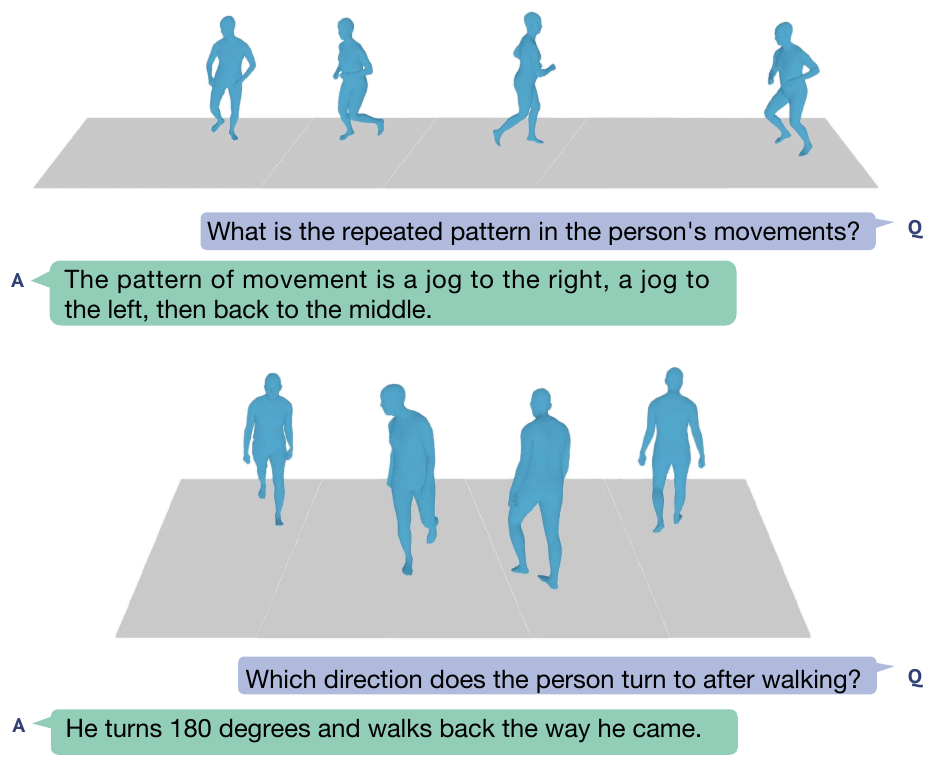}
    \caption{{\bf MoVid-Bench samples (direction awareness).} 
    }
    \label{fig:bench_example_dir}
\end{figure}

\begin{figure}[!t]
    \centering
    \includegraphics[width=0.7\textwidth]{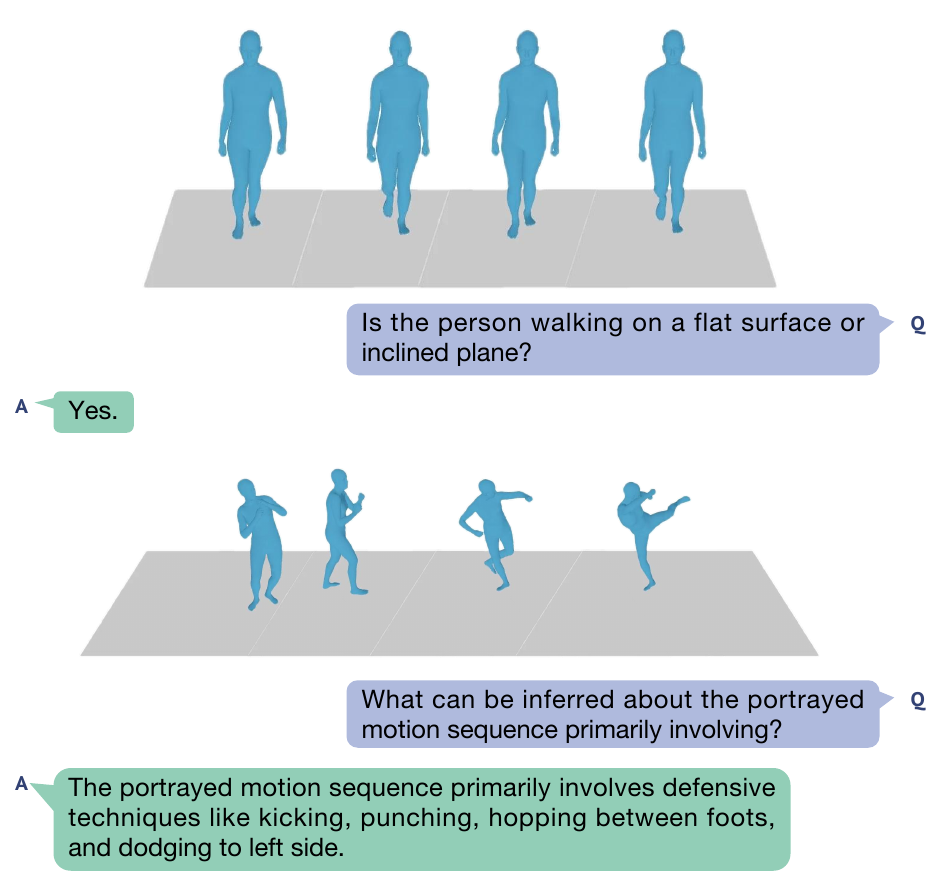}
    \caption{{\bf MoVid-Bench samples (reasoning ability).} 
    }
    \label{fig:bench_example_rea}
\end{figure}

\begin{figure}[!t]
    \centering
    \includegraphics[width=0.7\textwidth]{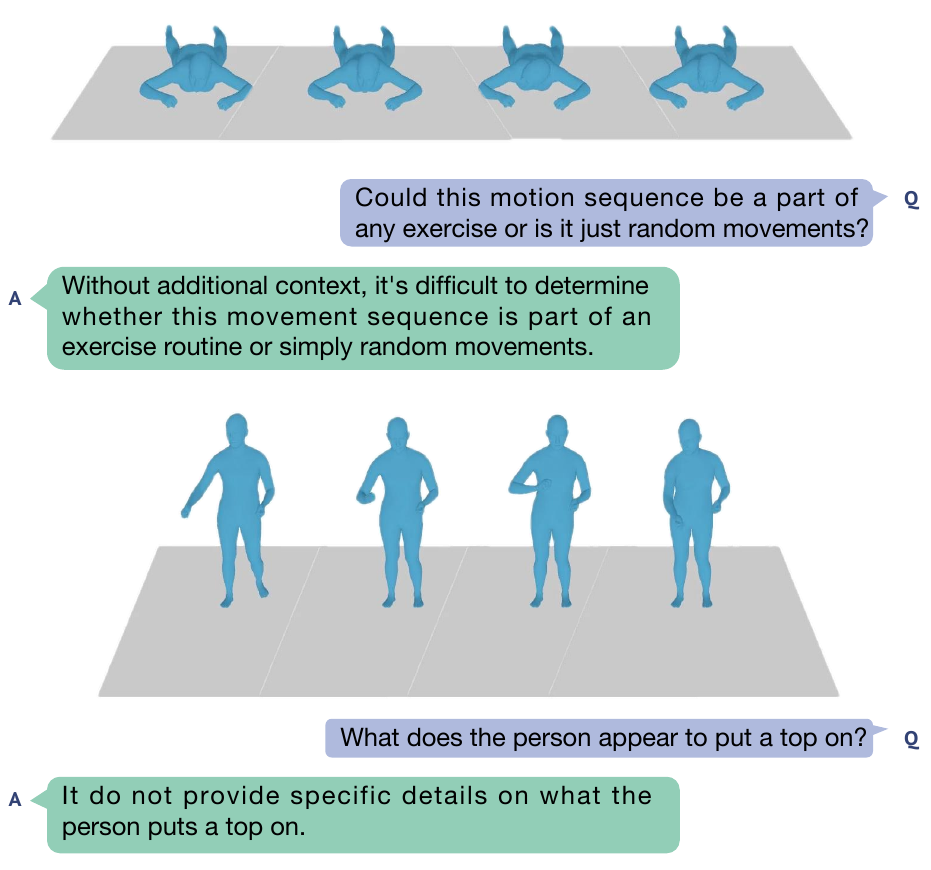}
    \caption{{\bf MoVid-Bench samples (robustness against hallucination).} 
    }
    \label{fig:bench_example_hall}
\end{figure}

Note that all examples have been manually annotated and checked carefully, which ensures the quality and fairness of our evaluation. We provide some examples of these aspects of motion part in~\cref{fig:bench_example_body}, \cref{fig:bench_example_seq}, \cref{fig:bench_example_dir}, \cref{fig:bench_example_rea}, and \cref{fig:bench_example_hall}, respectively. The video part of MoVid-Bench is designed similarly to the motion part. 

\newpage

\clearpage
\section{Visualization Comparison on Dataset Usage}

We compare the effect of data usage on both modalities via visualization results. To explore how motion and videos help each other to boost human behaviors, we show visualization comparisons besides numeric comparisons in the main paper. 

\vspace{-1em}
\subsection{Comparison on Whether to Use Motion Data}

We explore how motion data helps video understanding on human behaviors via visualization results. As shown in~\cref{tab:1_input_sequence} and~\cref{tab:2_input_sequence}, motion data alleviates the problem of model hallucination while describing the content of motions more accurately. This is mainly due to the dense information density of motion data.

\begin{table}[!h]
\centering
\begin{subtable}[t]{\linewidth}
\begin{minipage}{0.99\columnwidth}\vspace{0mm}    \centering
    \begin{tcolorbox} 
        \raggedright
        \small
        ${\texttt{\bf Human}:} $ describe the motion \\
        \PredSty{\texttt{<VIDEO>}}\\
        \includegraphics[width=0.23\linewidth]{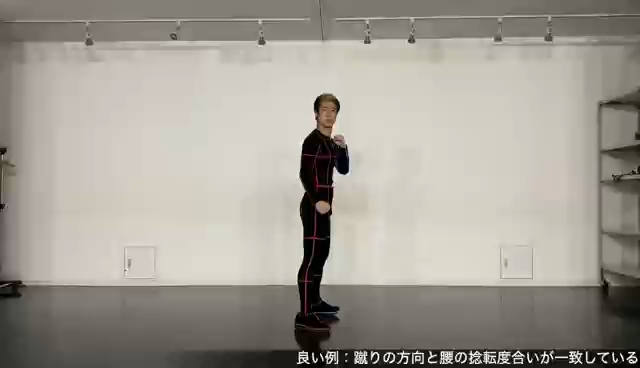}
        \includegraphics[width=0.23\linewidth]{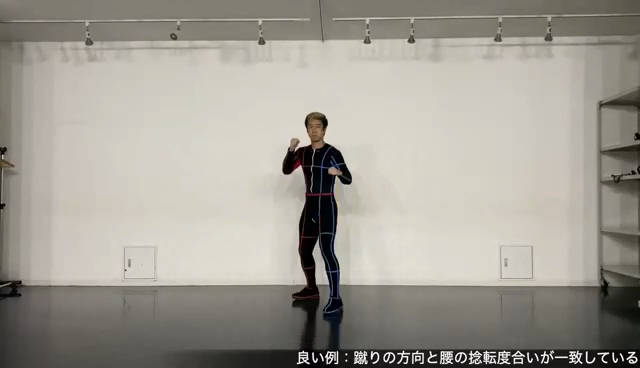}
        \includegraphics[width=0.23\linewidth]{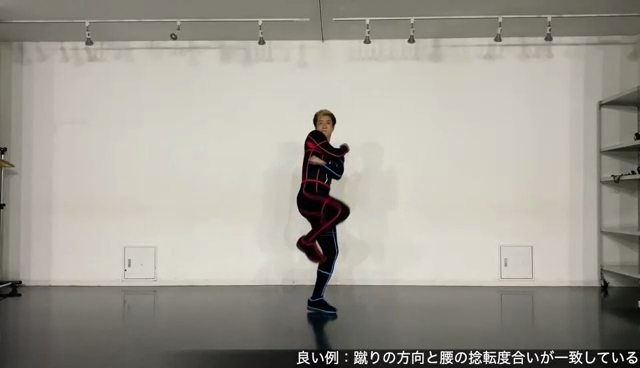}
        \includegraphics[width=0.23\linewidth]{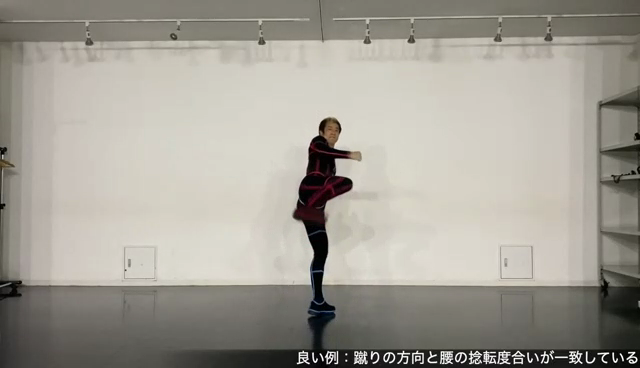}
        \includegraphics[width=0.23\linewidth]{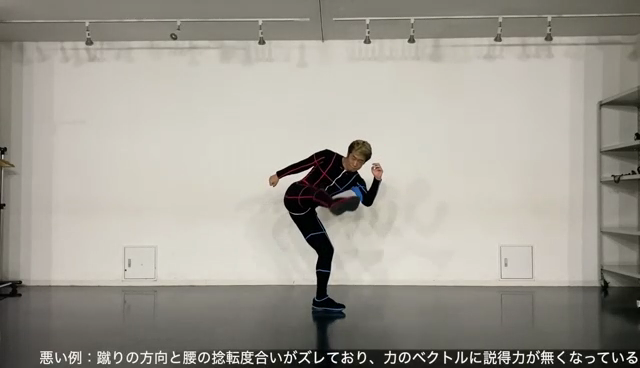}
        \includegraphics[width=0.23\linewidth]{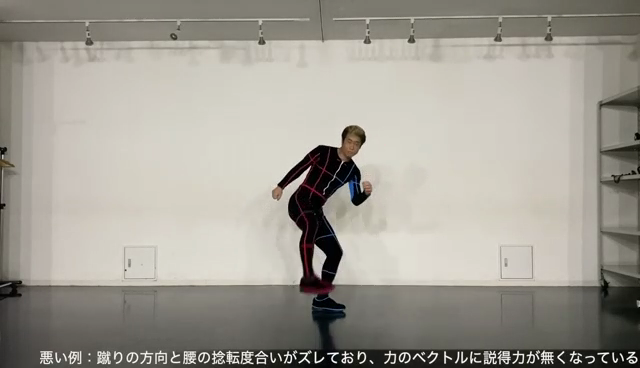}
        \includegraphics[width=0.23\linewidth]{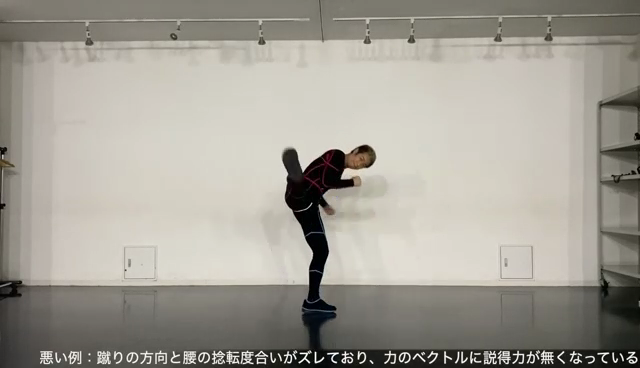}
        \includegraphics[width=0.23\linewidth]{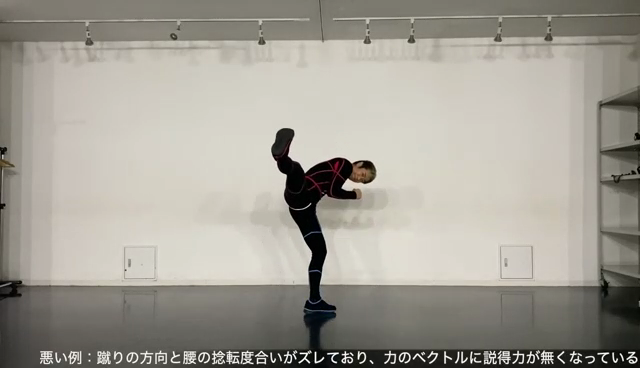}\\
        \PredSty{\texttt{<VIDEO>}}\\
        \smalllogo${\texttt{\bf MotionLLM}} (\texttt{w/o Motion data})$: The man in the video is seen moving his arms and legs around in a fluid motion, suggesting that he is performing a dance routine or a martial arts routine. 
    \end{tcolorbox}
    \caption{MotionLLM \textbf{w/o} Motion data training (video only).}
    \label{tab:input_sequence1}
\end{minipage}
\end{subtable}
\begin{subtable}[t]{\linewidth}
\begin{minipage}{0.99\columnwidth}\vspace{0mm}    \centering
    \begin{tcolorbox} 
        \raggedright
        \small
        ${\texttt{\bf Human}:} $ describe the motion \\
        \PredSty{\texttt{<VIDEO>}}\\
        \includegraphics[width=0.23\linewidth]{figure/ablation_imgs/1.png}
        \includegraphics[width=0.23\linewidth]{figure/ablation_imgs/10.png}
        \includegraphics[width=0.23\linewidth]{figure/ablation_imgs/38.png}
        \includegraphics[width=0.23\linewidth]{figure/ablation_imgs/46.png}
        \includegraphics[width=0.23\linewidth]{figure/ablation_imgs/169.png}
        \includegraphics[width=0.23\linewidth]{figure/ablation_imgs/174.png}
        \includegraphics[width=0.23\linewidth]{figure/ablation_imgs/192.png}
        \includegraphics[width=0.23\linewidth]{figure/ablation_imgs/197.png}\\
        \PredSty{\texttt{<VIDEO>}}\\
        \smalllogo${\texttt{\bf MotionLLM} (\texttt{w/o Motion data}):} $ The man is hitting recurring pads with his hands and feet, \textcolor{cvprblue}{kicking and throwing his bodys}. He jumps and swings his body back and forth. With each hit, the man seems to be \textcolor{cvprblue}{getting faster and more precise}. 
    \end{tcolorbox}
    \caption{MotionLLM \textbf{w/} Motion data training (joint training).}
    \label{tab:input_sequence2}
\end{minipage}
\end{subtable}
\caption{{\bf Comparison on whether using motion data.} The \textcolor{cvprblue}{blue} text highlights a comprehensive description of the motion and speed in the given video, a feature absent in training that lacks motion data. Training without motion data provides only a basic overview.}
\label{tab:1_input_sequence}
\end{table}

\begin{table}[!h]
\centering
\begin{subtable}[t]{\linewidth}
\begin{minipage}{0.99\columnwidth}\vspace{0mm}    \centering
    \begin{tcolorbox} 
        \raggedright
        \small
        ${\texttt{\bf Human}:} $ describe the body movement of the woman \\
        \PredSty{\texttt{<VIDEO>}}\\
        \includegraphics[width=0.23\linewidth]{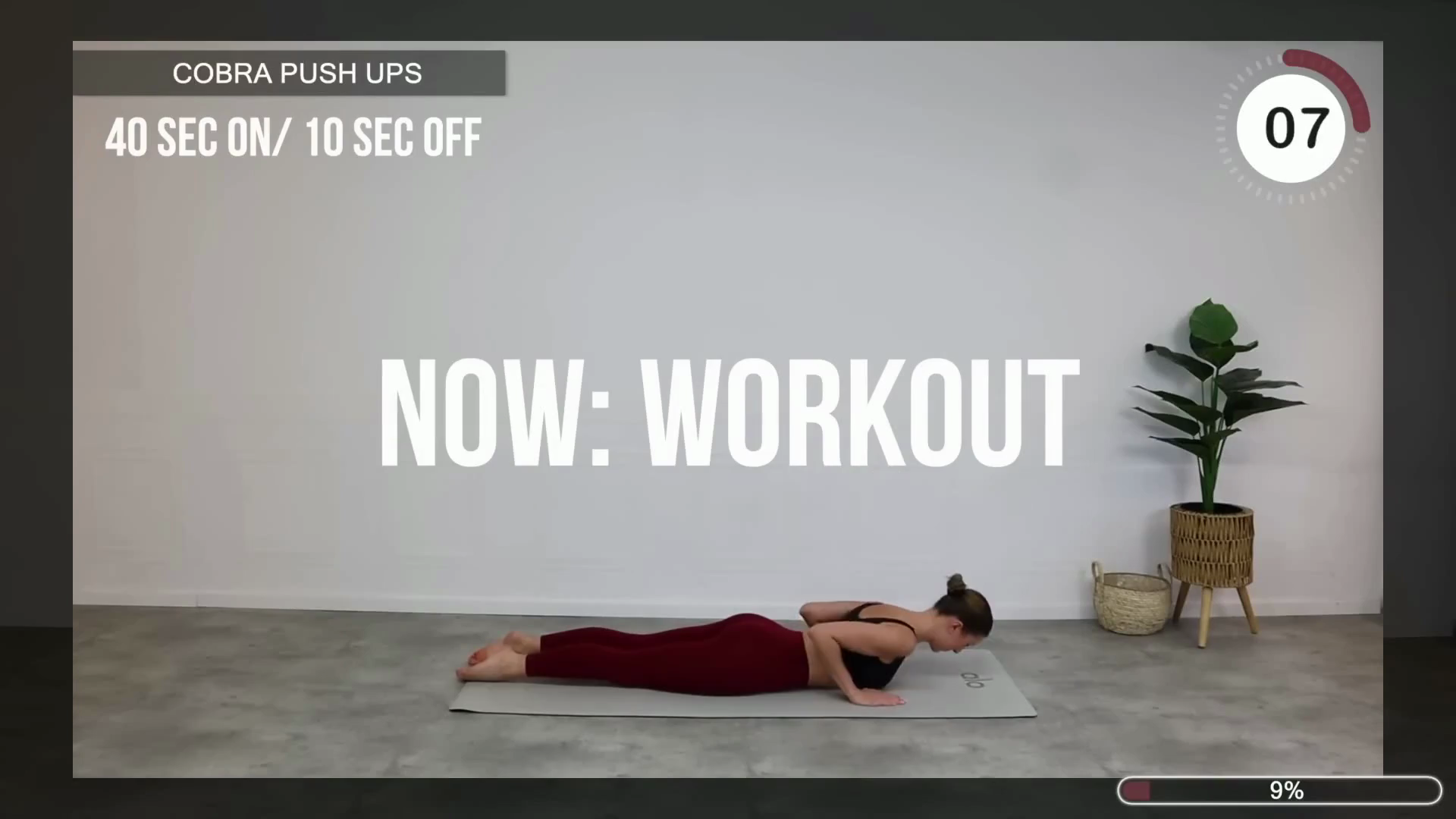}
        \includegraphics[width=0.23\linewidth]{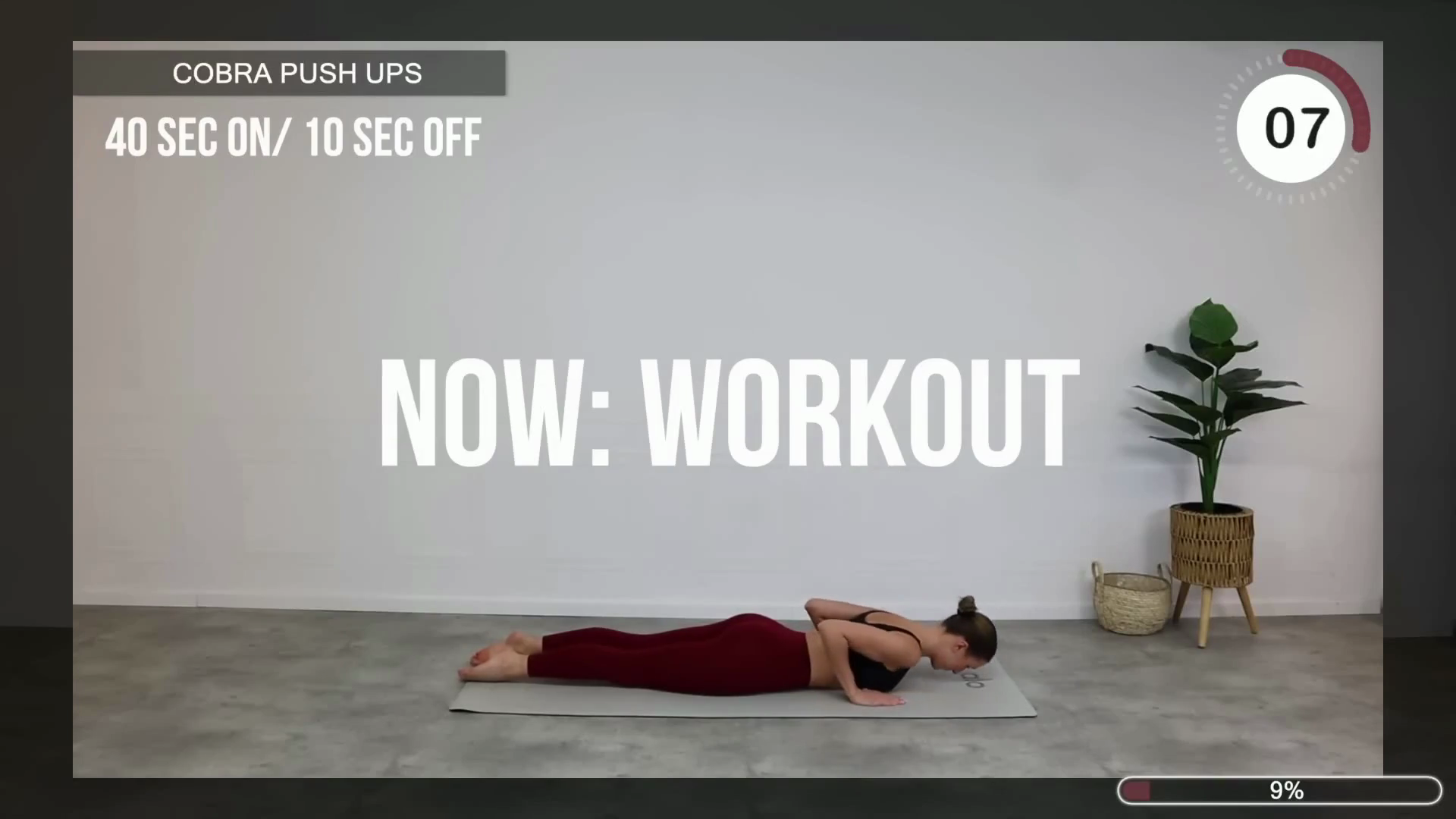}
        \includegraphics[width=0.23\linewidth]{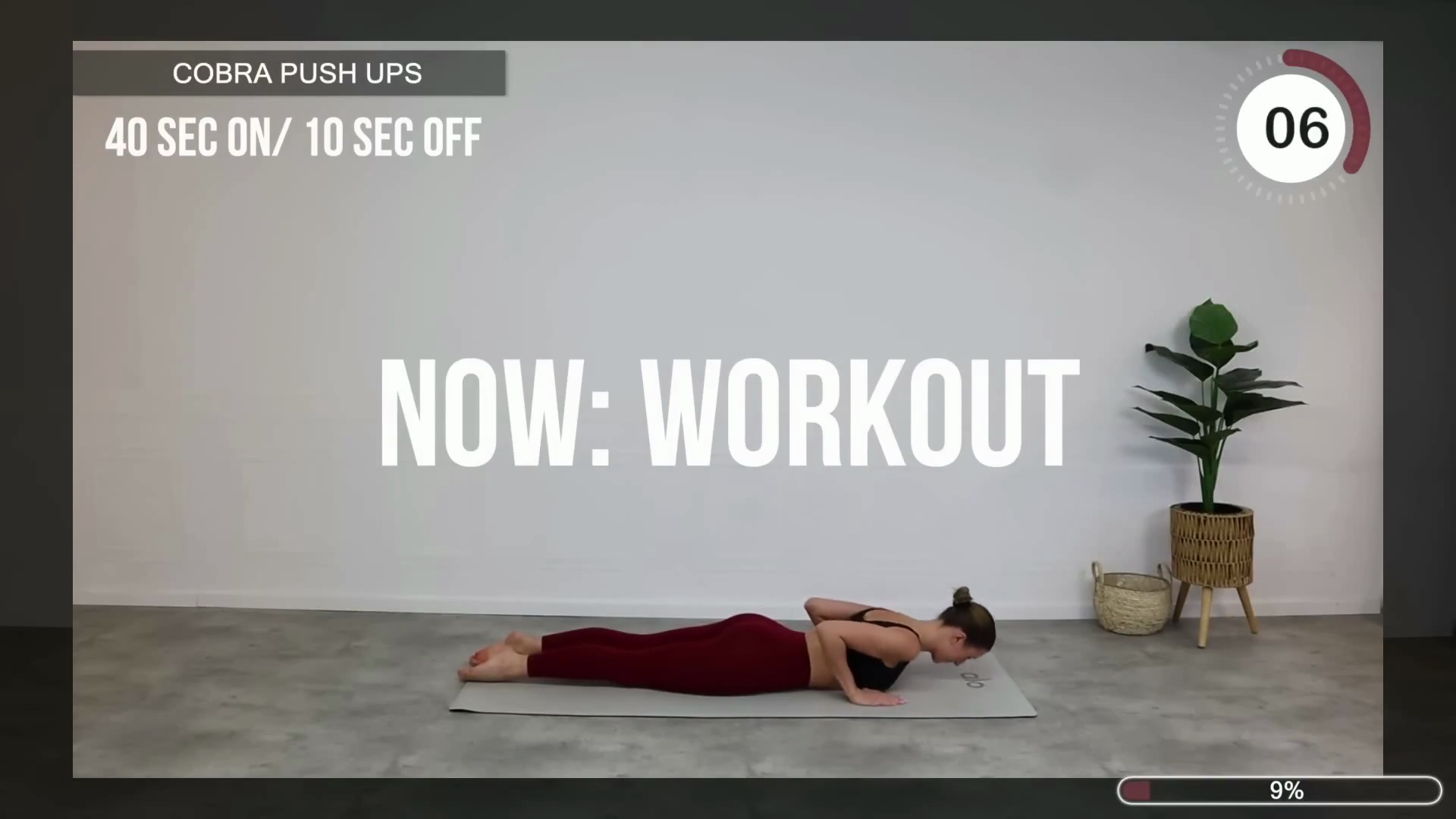}
        \includegraphics[width=0.23\linewidth]{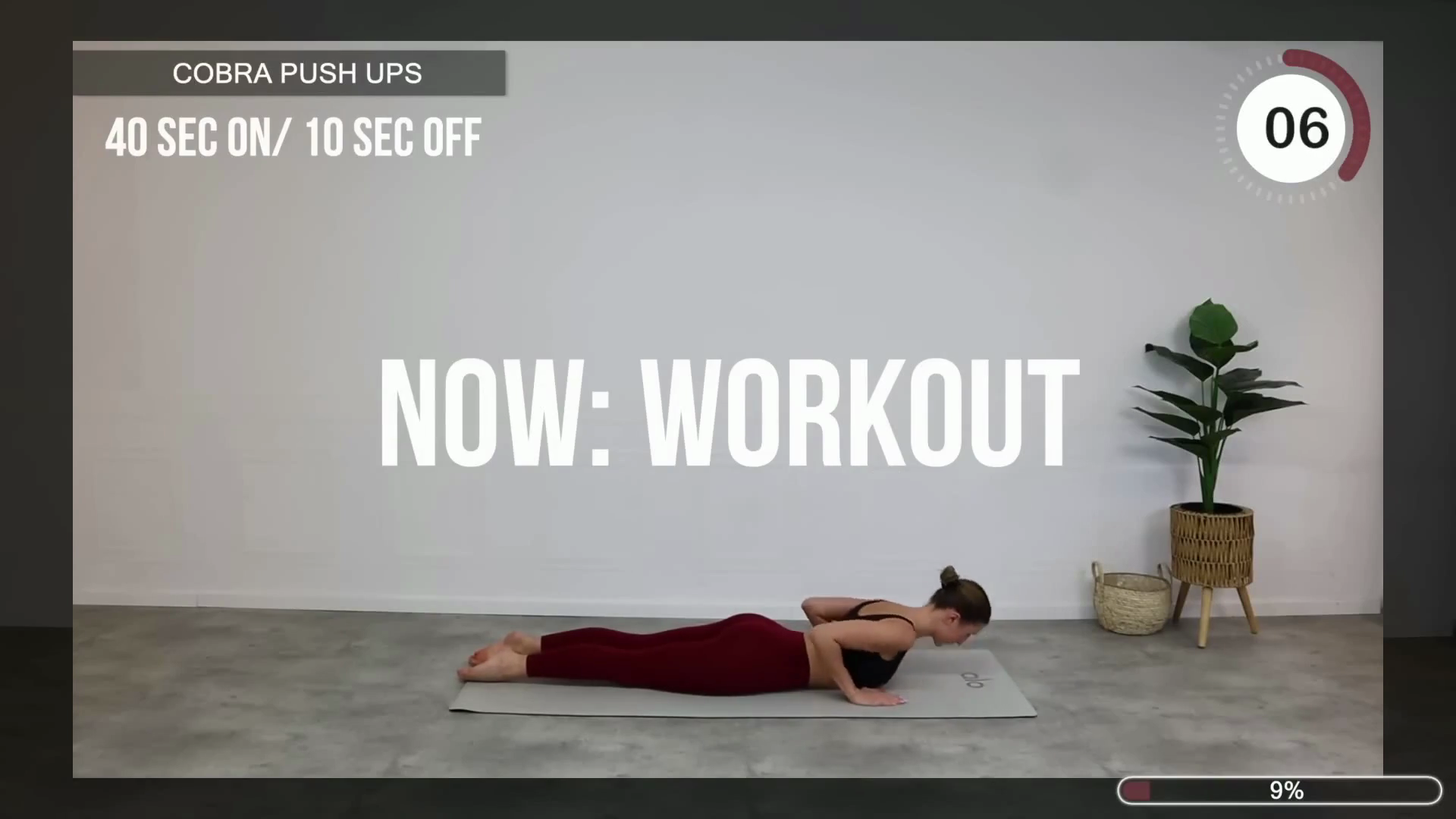}
        \includegraphics[width=0.23\linewidth]{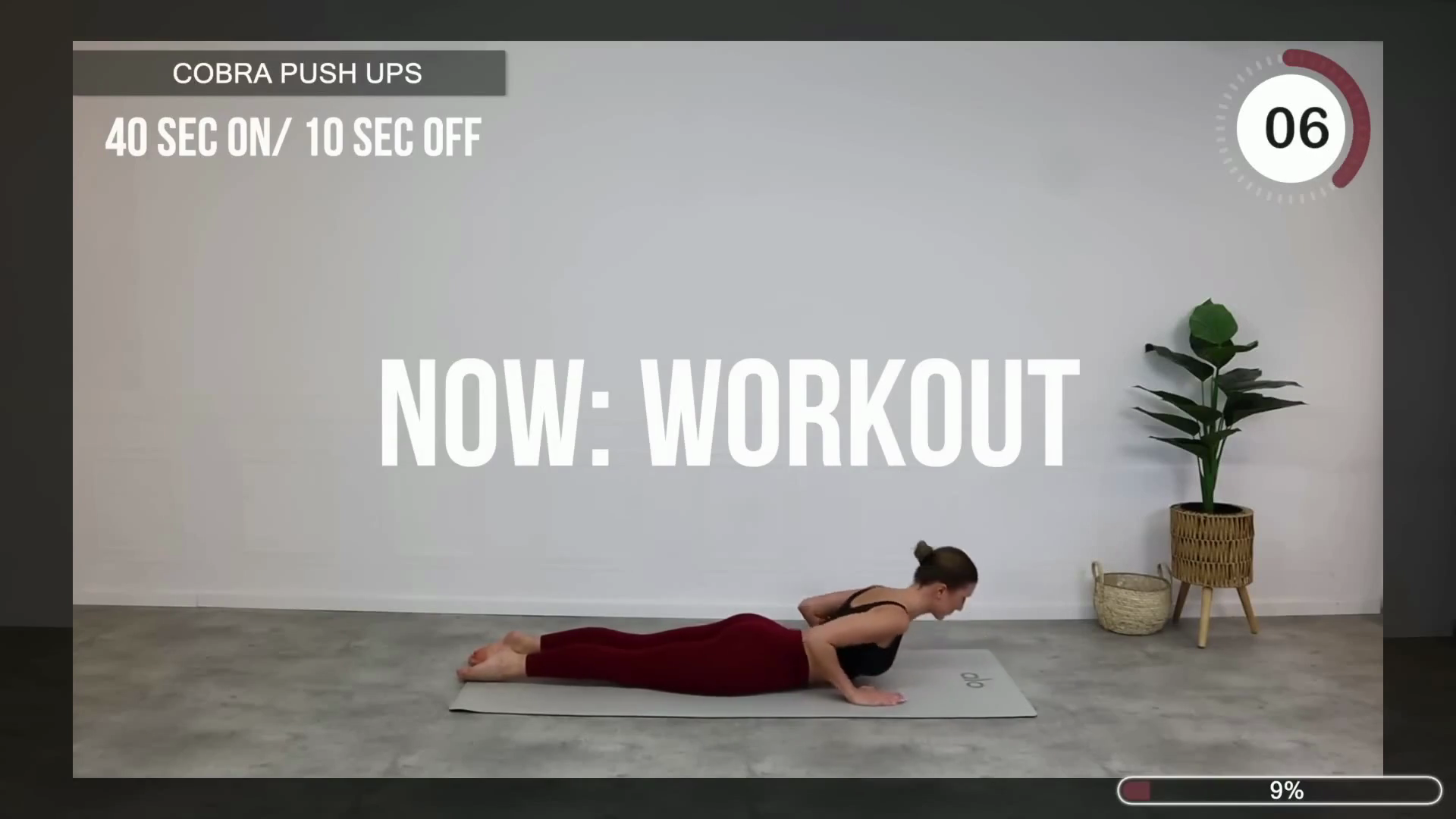}
        \includegraphics[width=0.23\linewidth]{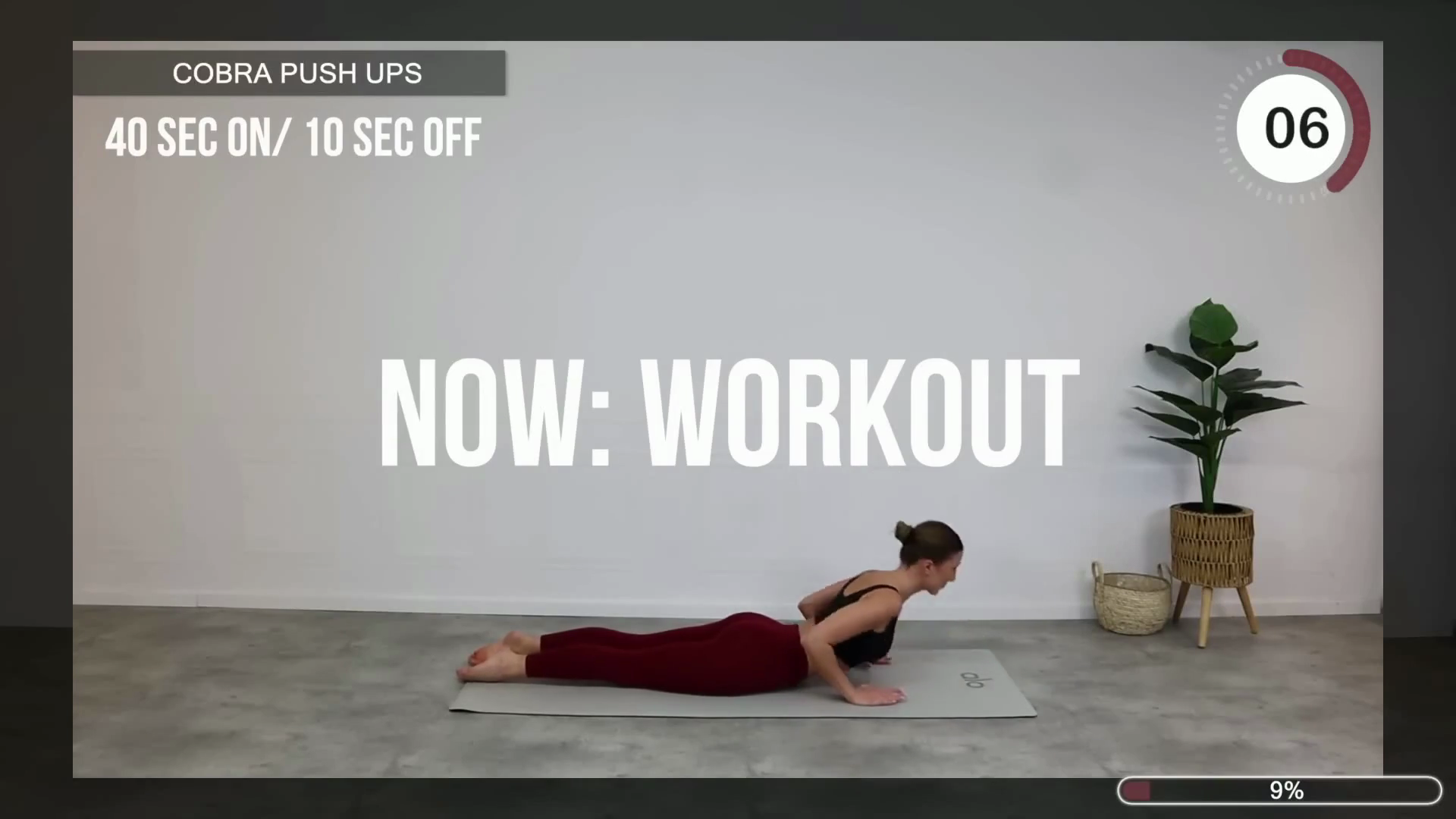}
        \includegraphics[width=0.23\linewidth]{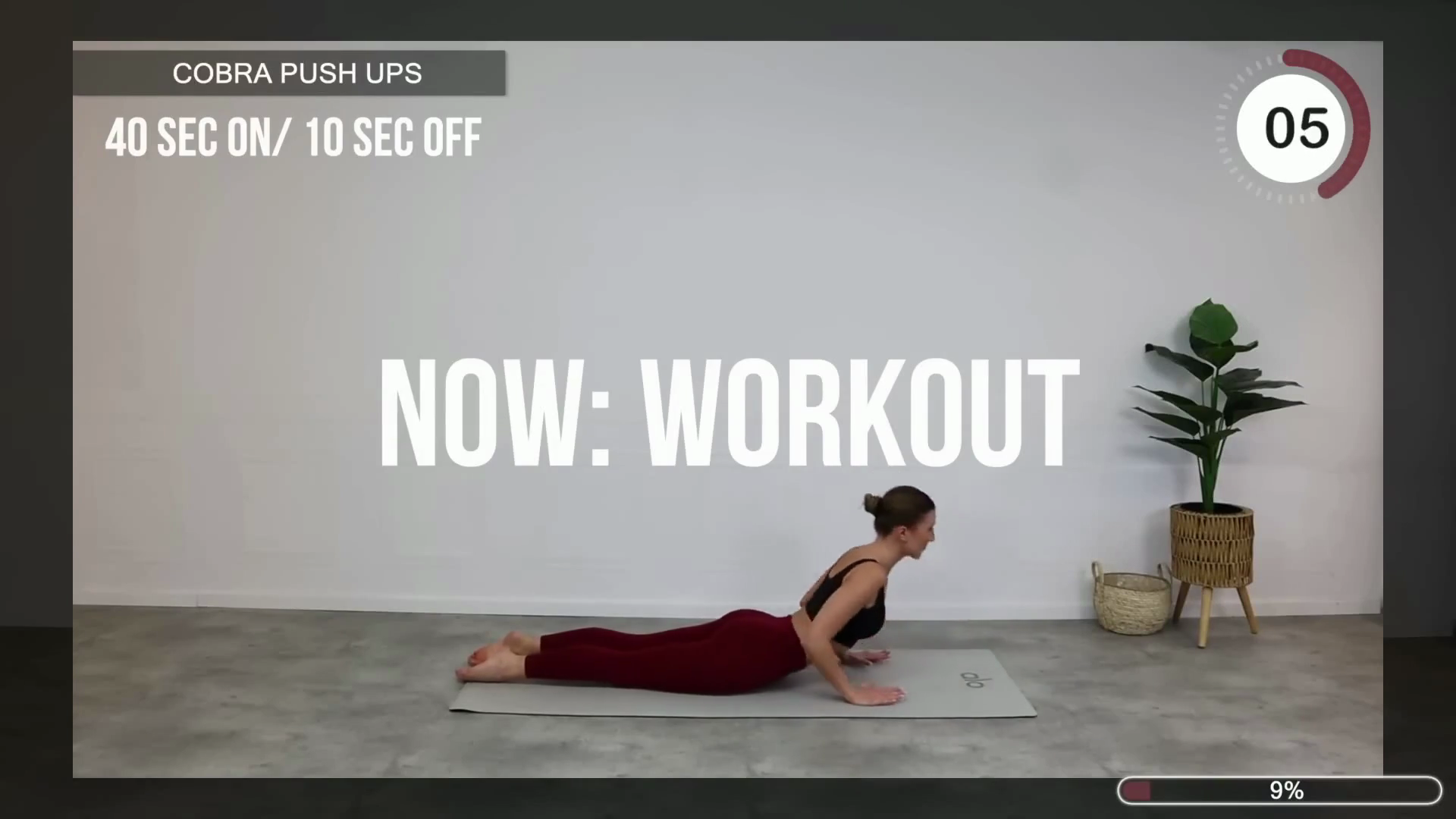}
        \includegraphics[width=0.23\linewidth]{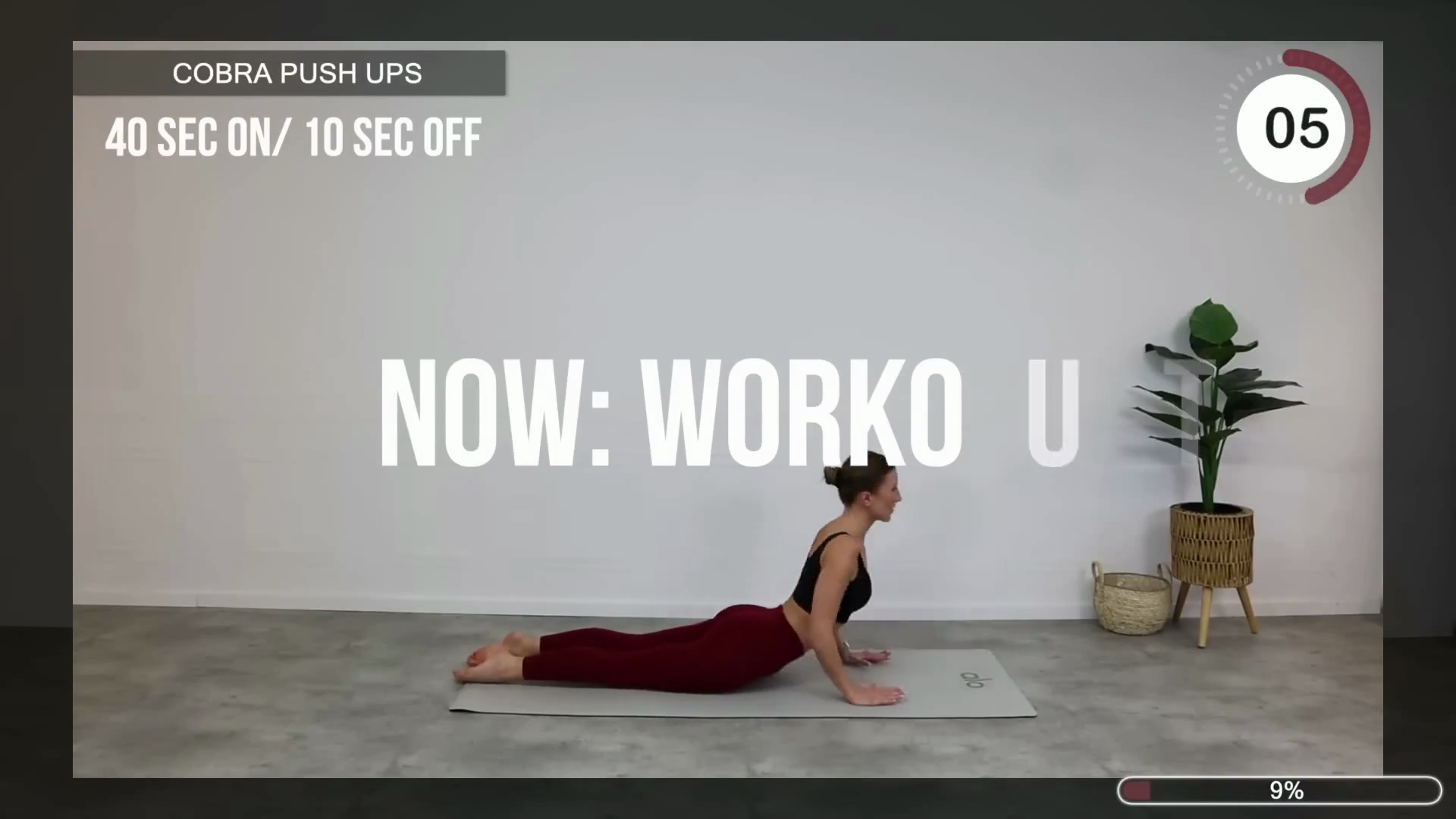}\\
        \PredSty{\texttt{<VIDEO>}}\\
        \smalllogo${\texttt{\bf MotionLLM}} (\texttt{w/o Motion data})$: The woman in the video performs a workout routine on a mat using her legs to support her body beneath her. \textcolor{pink}{She begins her routine by stretching her legs up the wall while lying on her back. She then proceeds to perform leg raises and leg sit-ups on her mat. After finishing her routine, the woman stops and stretches before walking away with a smile on her face.}
    \end{tcolorbox}
    \caption{MotionLLM w/o Motion data training (video only).}
    \label{tab:2_input_sequence1}
\end{minipage}
\end{subtable}
\begin{subtable}[t]{\linewidth}
\begin{minipage}{0.99\columnwidth}\vspace{0mm}    \centering
    \begin{tcolorbox} 
        \raggedright
        \small
        ${\texttt{\bf Human}:} $ describe the body movement of the woman \\
        \PredSty{\texttt{<VIDEO>}}\\
        \includegraphics[width=0.23\linewidth]{figure/ablation_imgs_2/1.png}
        \includegraphics[width=0.23\linewidth]{figure/ablation_imgs_2/12.png}
        \includegraphics[width=0.23\linewidth]{figure/ablation_imgs_2/25.png}
        \includegraphics[width=0.23\linewidth]{figure/ablation_imgs_2/30.png}
        \includegraphics[width=0.23\linewidth]{figure/ablation_imgs_2/35.png}
        \includegraphics[width=0.23\linewidth]{figure/ablation_imgs_2/39.png}
        \includegraphics[width=0.23\linewidth]{figure/ablation_imgs_2/45.png}
        \includegraphics[width=0.23\linewidth]{figure/ablation_imgs_2/53.png}\\
        \PredSty{\texttt{<VIDEO>}}\\
        \smalllogo${\texttt{\bf MotionLLM} (\texttt{w/ Motion data}):} $ The woman's body movements are slow and controlled, with her \textcolor{cvprblue}{lifting her upper body off the mat and her lower body following the movement. She maintains a straight back and neck throughout the movement.} 
    \end{tcolorbox}
    \caption{MotionLLM w/ Motion data training (joint training).}
    \label{tab:2_input_sequence2}
\end{minipage}
\end{subtable}
\caption{{\bf Comparison on whether using motion data for hallucinations.}The \textcolor{cvprblue}{blue} text specifically details the woman's bodily movements, whereas the \textcolor{pink}{pink} text merely describes hallucinations, which significantly differ from the actual content of the video.}
\label{tab:2_input_sequence}
\end{table}

\clearpage
\subsection{Comparison on Whether to Use Video Data}

We explore how video data helps motion understanding of human behaviors via visualization results. As shown in~\cref{tab:aba_motion_1} and~\cref{tab:aba_motion_2}, motion data can better reason about the environment and interaction content in which motions occur. This is mainly due to the fact that MotionLLM can obtain interactive motion grounds from a large number of videos.

\vspace{-1em}
\begin{table}[!h]
\centering
\begin{subtable}[t]{\linewidth}
\begin{minipage}{0.99\columnwidth}\vspace{0mm}    \centering
    \begin{tcolorbox} 
        \raggedright
        \small
        ${\texttt{\bf Human}:} $ Form a written description that correlates with the series of human motion tokens provided. \\
        \PredSty{\texttt{<MOTION>}}\\
        \includegraphics[width=0.23\linewidth]{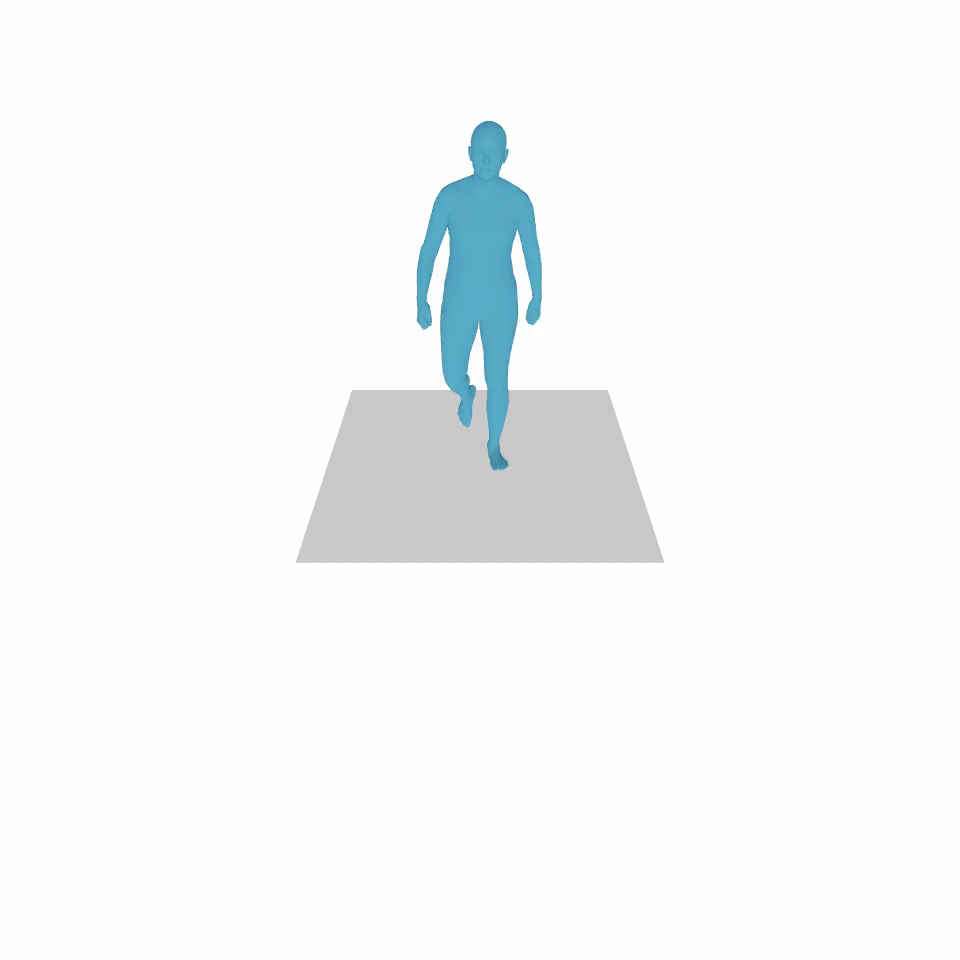}
        \includegraphics[width=0.23\linewidth]{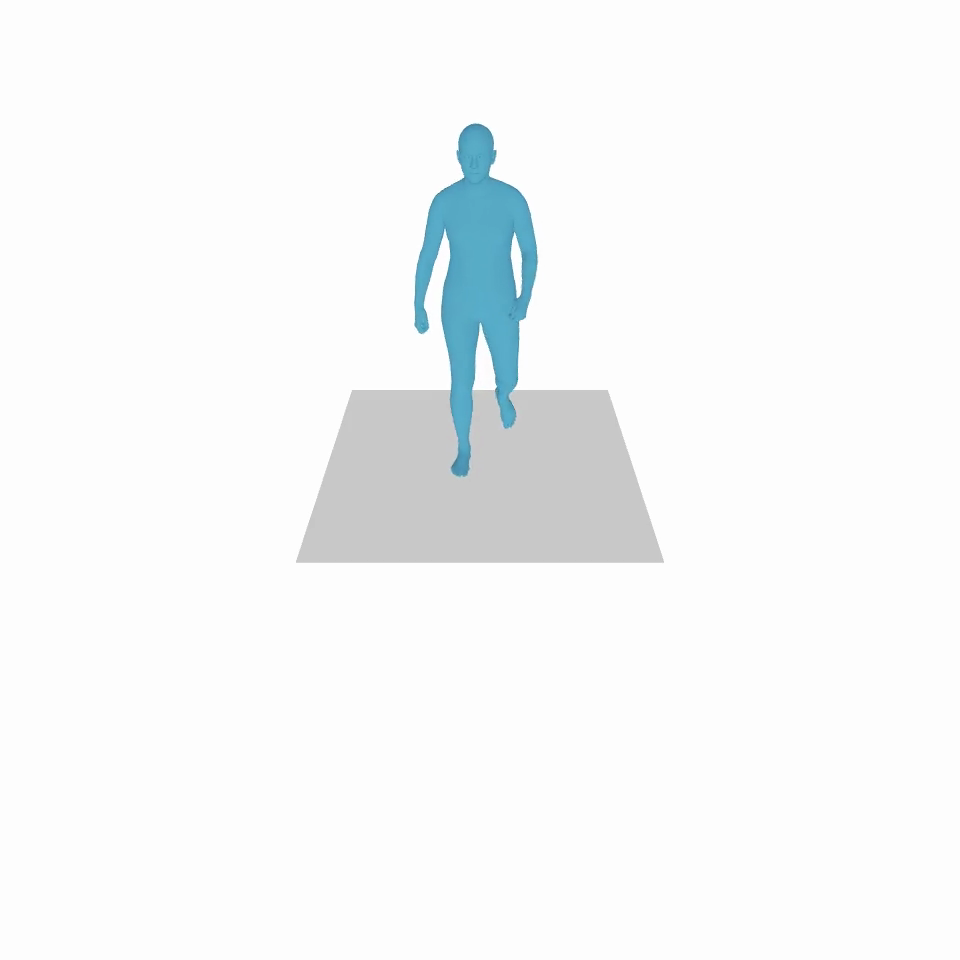}
        \includegraphics[width=0.23\linewidth]{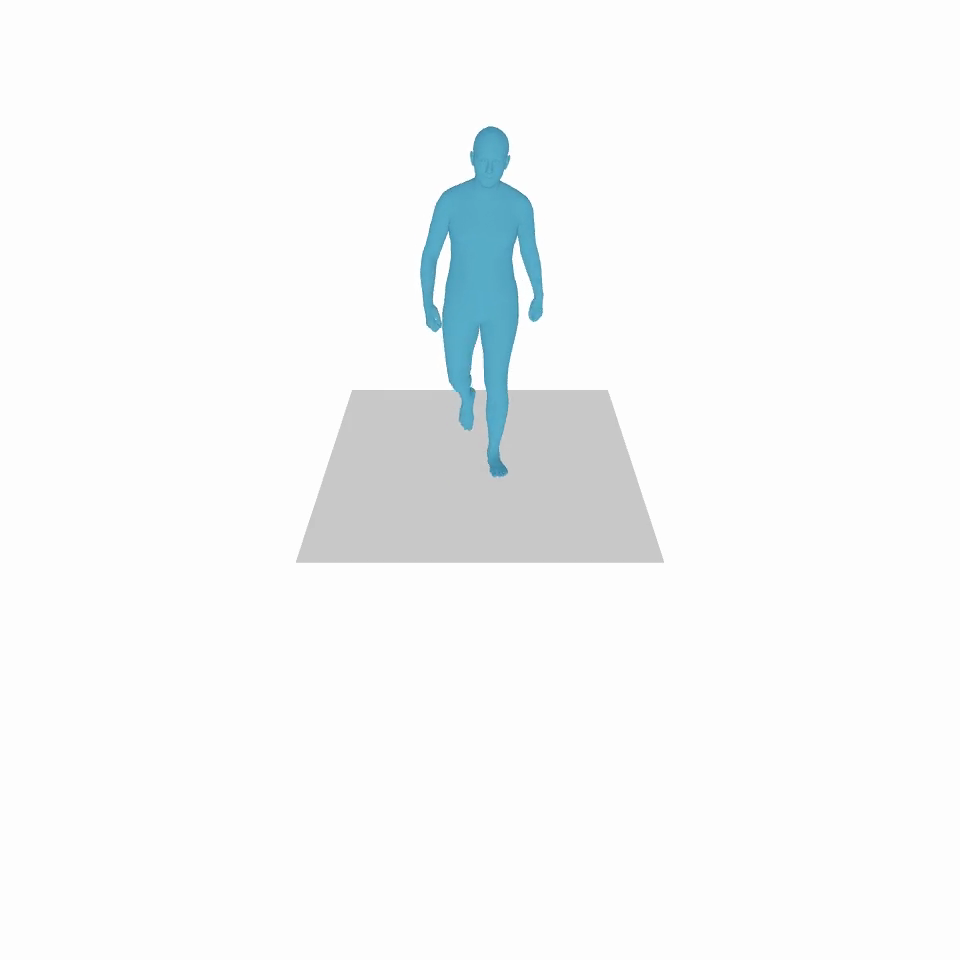}
        \includegraphics[width=0.23\linewidth]{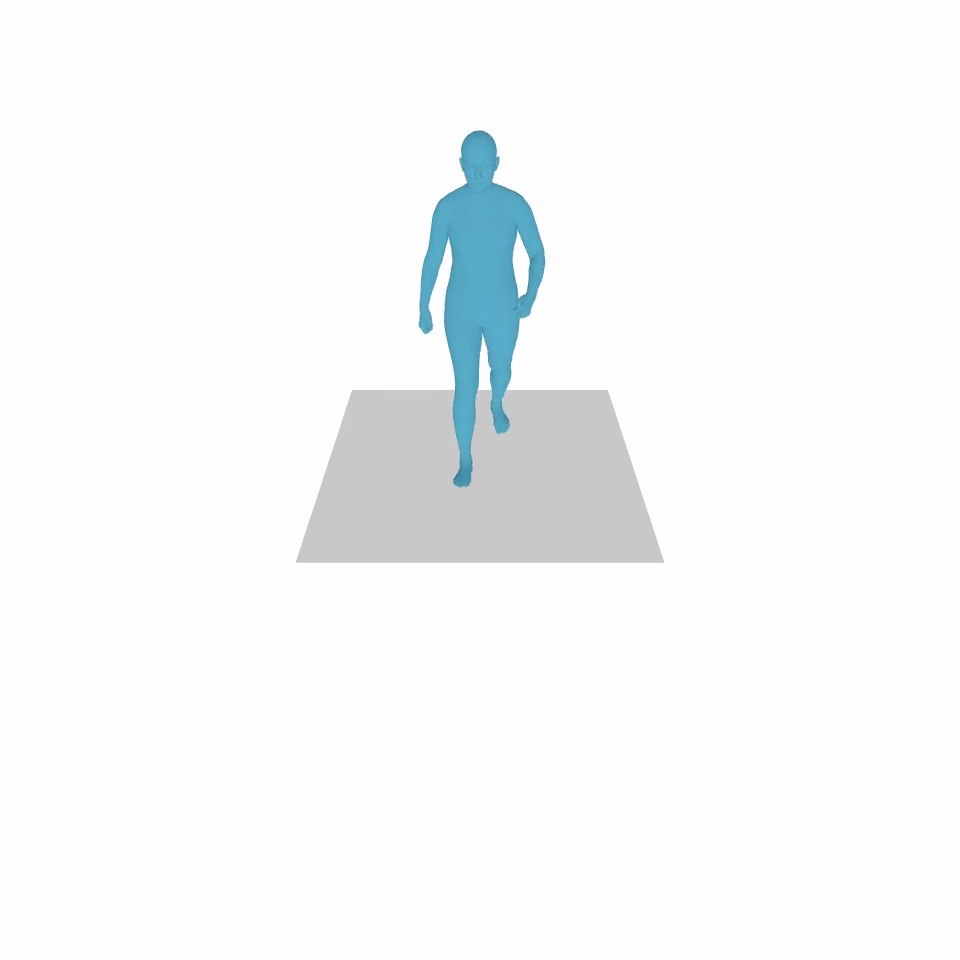}\\
        \PredSty{\texttt{<MOTION>}}\\
        \smalllogo${\texttt{\bf MotionLLM}} (\texttt{w/o Video data})$: The man is walking forward. \\
        ${\texttt{\bf Human}:} $ Where he might be walking?\\
        \smalllogo${\texttt{\bf MotionLLM}} (\texttt{w/o Video data})$: \textcolor{pink}{Forward}.
    \end{tcolorbox}
    \caption{MotionLLM w/o video data training (motion only).}
    \label{tab:aba_motion_1_1}
\end{minipage}
\end{subtable}
\begin{subtable}[t]{\linewidth}
\begin{minipage}{0.99\columnwidth}\vspace{0mm}    \centering
    \begin{tcolorbox} 
        \raggedright
        \small
        ${\texttt{\bf Human}:} $ Form a written description that correlates with the series of human motion tokens provided. \\
        \PredSty{\texttt{<MOTION>}}\\
        \includegraphics[width=0.23\linewidth]{figure/ablation_motion/case_1/1.png}
        \includegraphics[width=0.23\linewidth]{figure/ablation_motion/case_1/15.png}
        \includegraphics[width=0.23\linewidth]{figure/ablation_motion/case_1/29.png}
        \includegraphics[width=0.23\linewidth]{figure/ablation_motion/case_1/42.png}\\
        \PredSty{\texttt{<MOTION>}}\\
        \smalllogo${\texttt{\bf MotionLLM} (\texttt{w/o Video data}):} $ a man walking forward at the same place.\\
        ${\texttt{\bf Human}:} $ Where he might be walking?\\
        \smalllogo${\texttt{\bf MotionLLM} (\texttt{w/o Video data}):} $ \textcolor{cvprblue}{On a treadmill}. 
    \end{tcolorbox}
    \vspace{-0.5em}
    \caption{MotionLLM w/ video data training (joint training).}
    \label{tab:aba_motion_1_2}
\end{minipage}
\end{subtable}
\caption{{\bf Comparison on whether using video data.} Video data helps to infer the environment content of the ``treadmill'' due to the large number of video grounds in training.}
\label{tab:aba_motion_1}
\end{table}

\begin{table}[!h]
\centering
\begin{subtable}[t]{\linewidth}
\begin{minipage}{0.99\columnwidth}\vspace{0mm}    \centering
    \begin{tcolorbox} 
        \raggedright
        \small
        ${\texttt{\bf Human}:} $ Translate the given human motion into a corresponding textual description.\\
        \PredSty{\texttt{<MOTION>}}\\
        \includegraphics[width=0.23\linewidth]{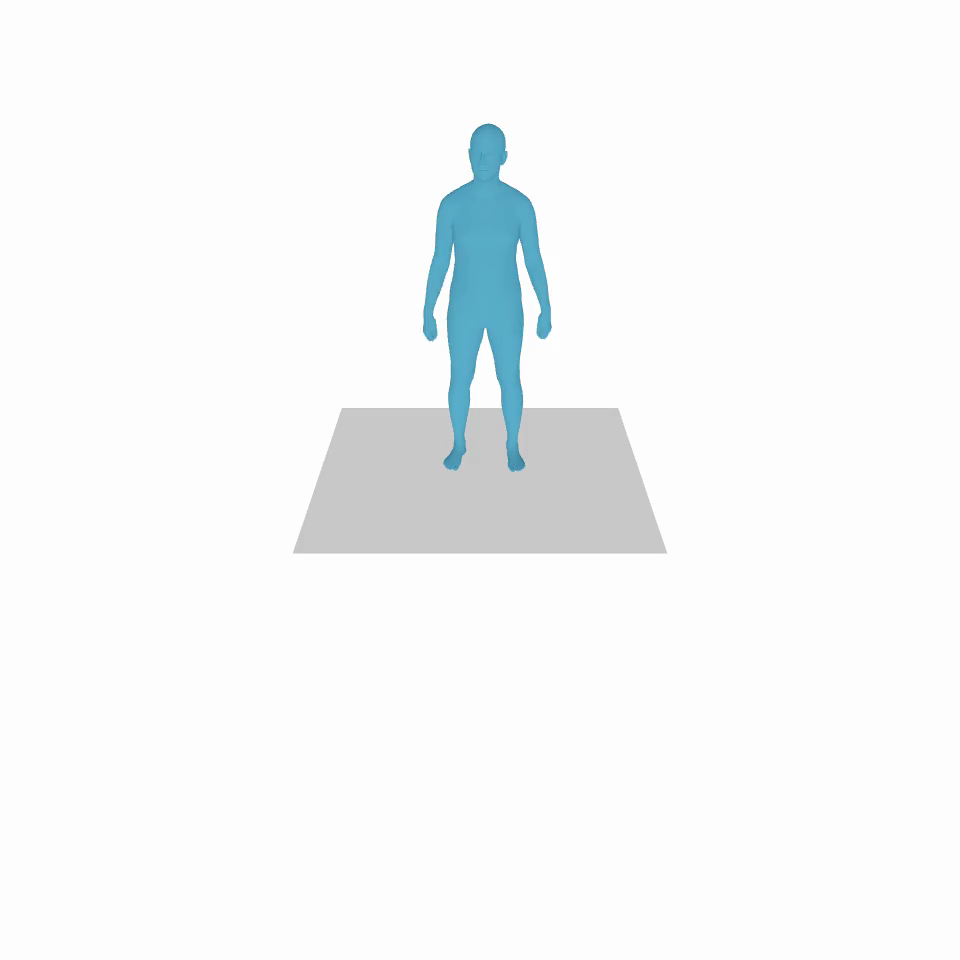}
        \includegraphics[width=0.23\linewidth]{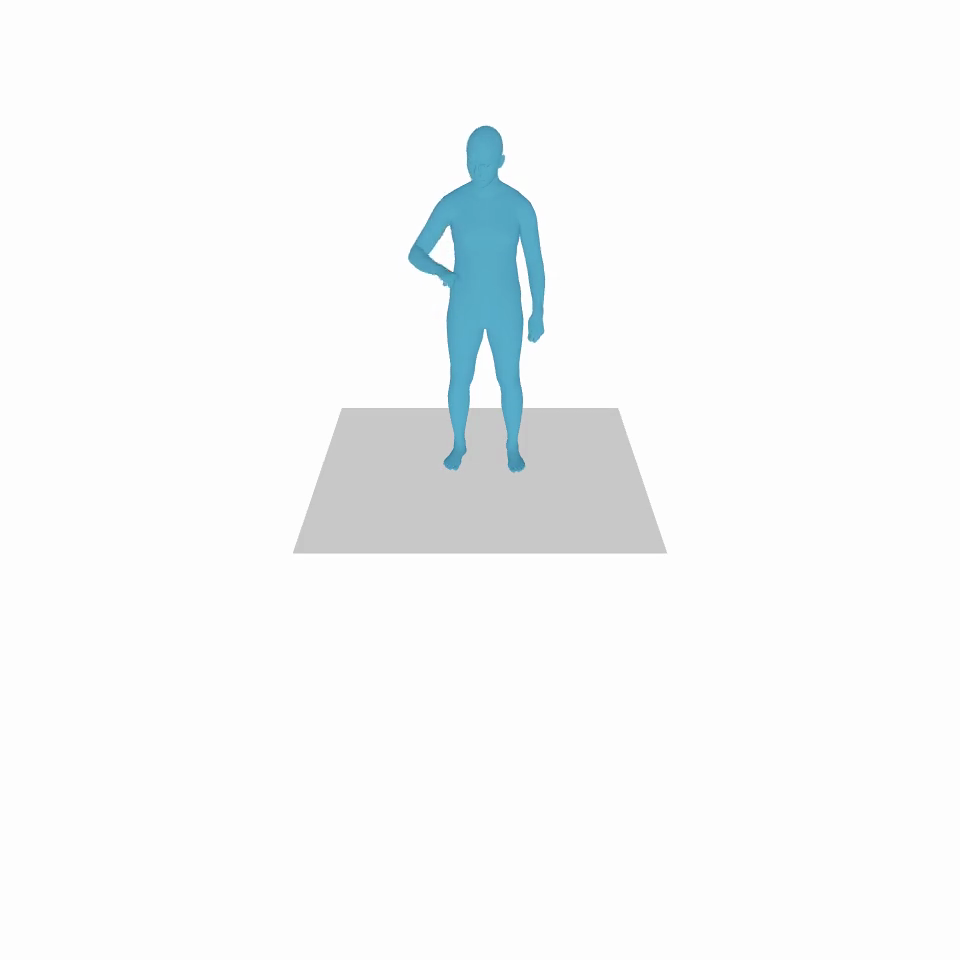}
        \includegraphics[width=0.23\linewidth]{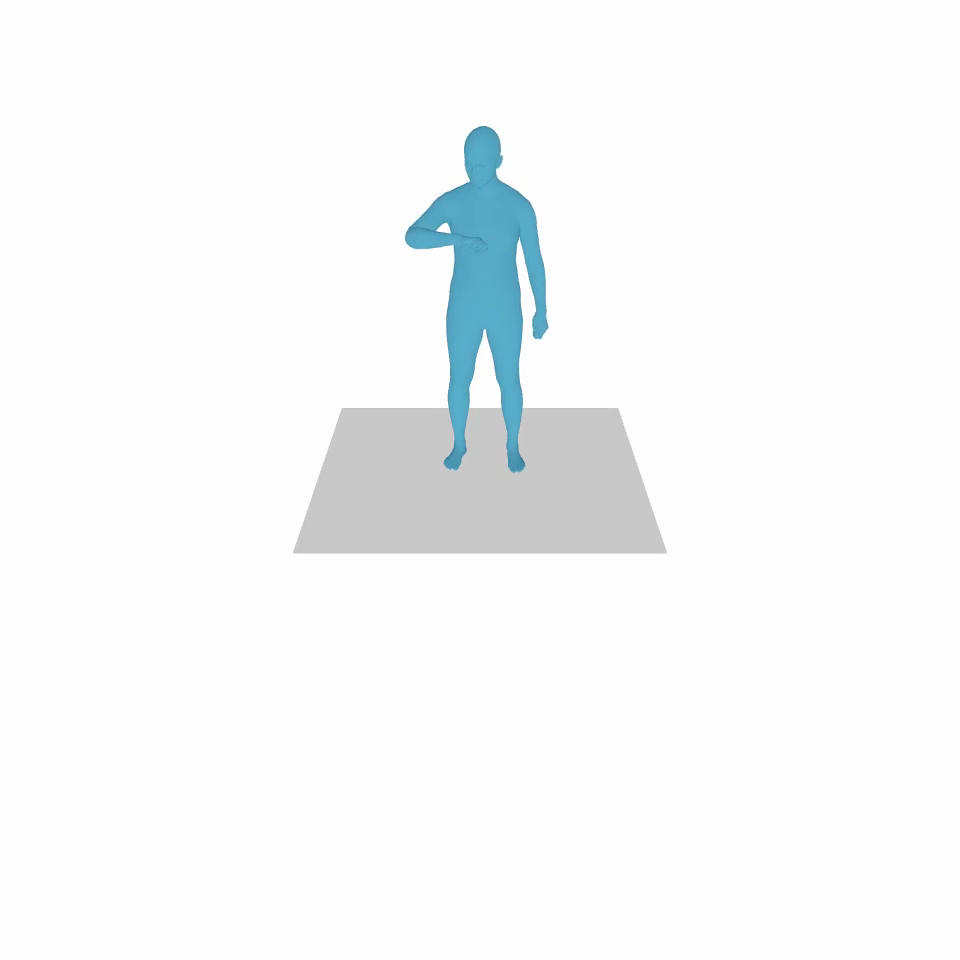}
        \includegraphics[width=0.23\linewidth]{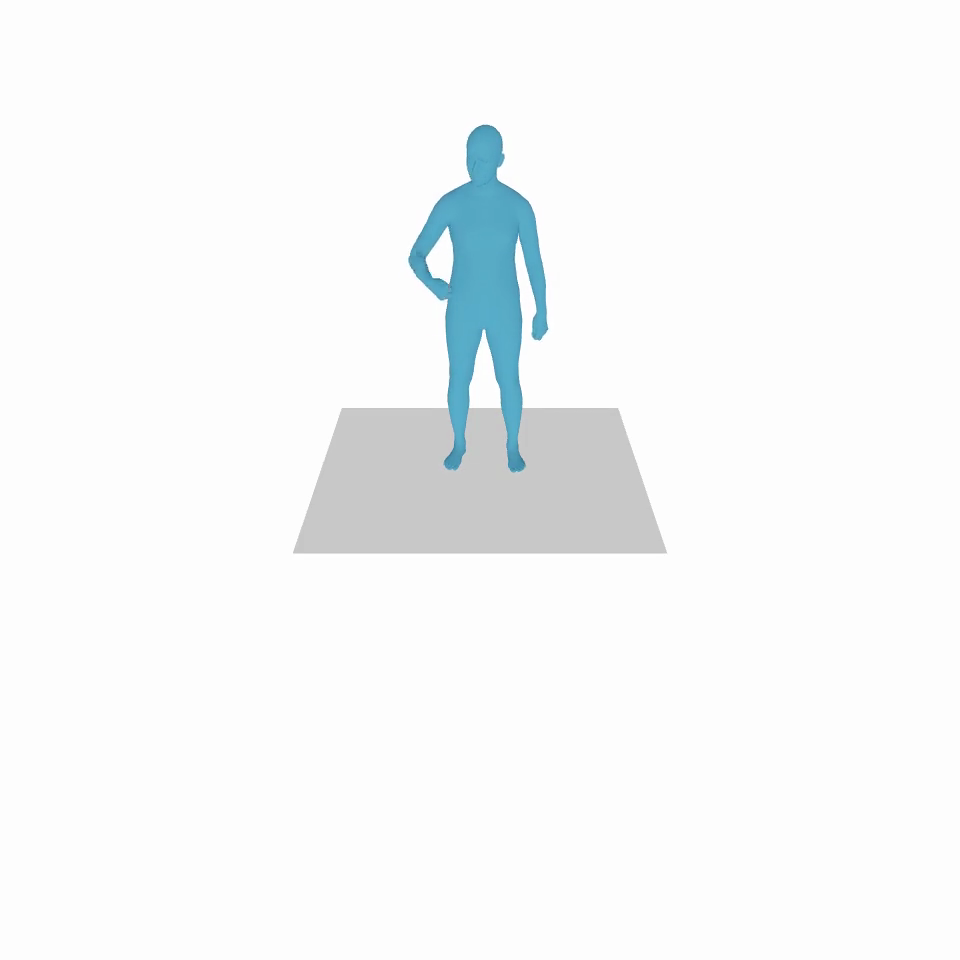}\\
        \PredSty{\texttt{<MOTION>}}\\
        \smalllogo${\texttt{\bf MotionLLM}} (\texttt{w/o Video data})$: the person stands there and raises their hands almost to their face.\\
        ${\texttt{\bf Human}:} $ What might be the motivation of it?\\
        \smalllogo${\texttt{\bf MotionLLM}} (\texttt{w/o Video data})$: It's \textcolor{pink}{hard to determine}. 
    \end{tcolorbox}
    \caption{MotionLLM w/o video data training (motion only).}
    \label{tab:aba_motion_2_1}
\end{minipage}
\end{subtable}
\begin{subtable}[t]{\linewidth}
\begin{minipage}{0.99\columnwidth}\vspace{0mm}    \centering
    \begin{tcolorbox} 
        \raggedright
        \small
        ${\texttt{\bf Human}:} $ Translate the given human motion into a corresponding textual description.\\
        \PredSty{\texttt{<MOTION>}}\\
        \includegraphics[width=0.23\linewidth]{figure/ablation_motion/case_2/1.png}
        \includegraphics[width=0.23\linewidth]{figure/ablation_motion/case_2/14.png}
        \includegraphics[width=0.23\linewidth]{figure/ablation_motion/case_2/31.png}
        \includegraphics[width=0.23\linewidth]{figure/ablation_motion/case_2/50.png}\\
        \PredSty{\texttt{<MOTION>}}\\
        \smalllogo${\texttt{\bf MotionLLM} (\texttt{w/o Video data}):} $ The man raises the right hand.\\
        ${\texttt{\bf Human}:} $ What might be the motivation of it?\\
        \smalllogo${\texttt{\bf MotionLLM} (\texttt{w/o Video data}):} $ It might be looking at \textcolor{cvprblue}{the watch on the right wrist}.
    \end{tcolorbox}
    \caption{MotionLLM w/ video data training (joint training).}
    \label{tab:aba_motion_2_2}
\end{minipage}
\end{subtable}
\caption{{\bf Comparison on whether using video data.} The second dialogue helps the comprehension and reasoning of the content of ``looking watch on the right wrist'', which is more accurate.}
\label{tab:aba_motion_2}
\end{table}

\end{document}